\documentclass[10pt,twocolumn,letterpaper]{article}

\usepackage[pagenumbers]{cvpr} 

\definecolor{cvprblue}{rgb}{0.21,0.49,0.74}
\usepackage[pagebackref,breaklinks,colorlinks,allcolors=cvprblue]{hyperref}
\usepackage{url}
\usepackage{changepage}
\usepackage{graphicx}
\usepackage{makecell}
\usepackage{enumitem}
\usepackage{colortbl}
\usepackage{tikz}
\usepackage{amsmath}
\usepackage{xcolor}
\usepackage{tcolorbox}
\usepackage{multicol}
\usepackage{multirow}

\definecolor{mygreen}{HTML}{3cb44b}
\newcommand{\VarSty}[1]{\textnormal{\ttfamily\color{blue!90!black}#1}\unskip}

\def\paperID{7970} 
\def\confName{CVPR}
\def\confYear{2025}

\title{LLaVA-Critic: Learning to Evaluate Multimodal Models}

\author{Tianyi Xiong$^{{\natural},\heartsuit}$, Xiyao Wang$^{{\natural},\heartsuit}$, Dong Guo$^{\dagger}$, Qinghao Ye$^{\dagger}$, Haoqi Fan$^{\dagger}$, Quanquan Gu$^{\dagger}$,\\
Heng Huang$^{{\natural}}$, Chunyuan Li$^{\dagger}$ \\
$^{\dagger}$ByteDance,~~ $^{\natural}$University of Maryland, College Park \vspace{1mm}\\
\href{https://llava-vl.github.io/blog/2024-10-03-llava-critic}{https://llava-vl.github.io/blog/llava-critic}
}

\newcommand{\ours}{LLaVA-Critic\xspace}

\begin{document}
\maketitle
\renewcommand{\thefootnote}{\fnsymbol{footnote}}
\footnotetext{$\heartsuit$ Work collaborated with ByteDance}
\renewcommand{\thefootnote}{\arabic{footnote}}
\begin{abstract}
We introduce LLaVA-Critic, the first open-source large multimodal model (LMM) designed as a generalist evaluator to assess performance across a wide range of multimodal tasks. LLaVA-Critic is trained using a high-quality critic instruction-following dataset that incorporates diverse evaluation criteria and scenarios. Our experiments demonstrate the model's effectiveness in two key areas: $(i)$ LMM-as-a-Judge, where LLaVA-Critic provides reliable evaluation scores, performing on par with or surpassing GPT models on multiple evaluation benchmarks; and $(ii)$ Preference Learning, where it generates reward signals for preference learning, enhancing model alignment capabilities. This work underscores the potential of open-source LMMs in self-critique and evaluation, setting the stage for future research into scalable, superhuman alignment feedback mechanisms for LMMs. 
\end{abstract}    
\section{Introduction}
\label{sec:intro}

The ability of learning to evaluate is increasingly taking on a pivotal role in the development of modern large multimodal models (LMMs), as pre-training on existing web data reaches maturity and the focus is shifting towards post-training with AI-enhanced synthetic data, which shows growing potential. Reliable AI evaluation is essential, not only for offering scalable solutions to reduce human labor in complex task assessments, but also for generating effective reward signals in reinforcement learning and guiding inference-time search~\citep{ouyang2022training,o1,snell2024scaling}.

Existing LMMs have made tremendous progress in recent research community~\citep{li2023multimodal}, primarily on improving the performance of various real-world vision tasks in single-image~\citep{liu2023llava,bai2023qwen,chen2023internvl}, multi-image~\citep{li2024llavanext-interleave, jiang2024mantis} and video scenarios~\citep{li2024llavaonevision,lin2024vila,wang2024qwen2}. It remains unexplored to develop open LMMs to play the role of a judge and evaluate the performance of multimodal models. 
For instance, a model can follow a well-designed, itemized evaluation criterion to provide a score between 1 and 10 for rating different model responses in a visual chat task~\citep{liu2023llava}. Along with the score, it would also offer the associated reasoning behind the evaluation, ensuring transparency and consistency in assessing model performance.
In this paper, we present the first attempt to curate the instruction-following data particularly for evaluation, based on which we develop a LMM, \ours. 
Two primary scenarios/goals of building \ours are highlighted: 

\begin{itemize}[leftmargin=2.5mm]
\setlength{\itemsep}{2pt}
\item {\it Scenario 1: LMM-as-a-Judge}. Open-source LMMs that can deliver reliable evaluation scores, comparable to or surpassing proprietary models like GPT-4V~\citep{openai2023gpt4v}/-4o~\citep{openai2024gpt4o}. These models offer a free alternative to replace commercial GPT models in various evaluation benchmarks.

\item {\it Scenario 2: Preference Learning}. A scalable solution for generating effective reward signals, reducing the need for costly human feedback collection. This approach enhances preference alignment with AI-generated feedback.
\end{itemize}

Our experimental results demonstrate that:
$(i)$ As a judge model, the evaluation scores and rankings provided by \ours show a high correlation with commercial GPT models, making it a cost-effective alternative for model developers in resource-constrained settings;
$(ii)$ In preference learning, \ours offers AI-generated feedback in iterative Direct Preference Optimization (DPO)~\citep{rafailov2024direct}, outperforming the preference signals provided by the reward model in LLaVA-RLHF~\citep{sun2023aligning}, which relies on human feedback for training the reward model.

\vspace{1mm}
In summary, our contributions are as follows:
\begin{itemize}[leftmargin=2.5mm]
\setlength{\itemsep}{2pt}
\item {\it Critic Instruction-Following Data}: We present a high-quality dataset tailored to follow instructions in complex evaluation setting to provide quantitative judgment and the corresponding reasoning process. It consists of 46k images with 113k evaluation instruction samples, including both pointwise and pairwise evaluation settings.

\item {\it Large Multimodal Models}: We develop \ours, LMMs that expand the capabilities of open models to act as critic, providing effective evaluation and feedback.

\item {\it Open-Source}: In an effort to support the development of general-purpose visual assistants, we release our critic instruction data, codebase, model checkpoints, and the trained visual chat demo to the public.
\end{itemize}

\section{Related Work}

\paragraph{LMM-as-a-judge.} 
Strong proprietary LMMs such as GPT-4V/-4o have been demonstrated to serve as generalist evaluators for vision-language tasks~\citep{zhang2023gpt,ge2023mllm}. Specifically, for complex scenarios related to human judgment, such as visual chat and detailed captioning, LMMs have been utilized in evaluation benchmarks to judge the model responses, in both pointwise~\citep{liu2023llava,yu2023mm,sun2023aligning,zhang2024lmmseval,li2024llavanext-strong,zhang2024mathverse} and pairwise settings~\citep{lu2024wildvision,yu2024rlaif}. 
Our LLaVA-Critic are employed in the these evaluation scenarios as open-source alternative, with advantages in cheap and customized evaluation.
For open-source models, Prometheus-Vision~\citep{lee2024prometheusvision} is the first VLM trained as an evaluator for specific user-designed scoring criteria. While sharing the same open-source spirit, LLaVA-Critic stands out as the first open generalist evaluator.
Note that GPT is also utilized to extract answers from LMM responses for subsequent evaluation in some benchmarks~\citep{lu2023mathvista, guan2024hallusionbench, wang-etal-2024-mementos}. This extractive functionality for evaluation is out of the scope of this paper.

\vspace{-3mm}
\paragraph{Preference learning for LMMs.} Reinforcement learning from human feedback (RLHF) is a proven method to align large language models (LLMs) with human intentions. DPO~\citep{rafailov2024direct} introduces a new parameterization of the reward model in RLHF, enabling direct optimization using pairwise preference datasets. 
CriticGPT~\citep{mcaleese2024llm} trains ``critic'' models to evaluate model-generated code, providing feedback signals to enhance code LLMs.
The concept of preference learning has recently expanded from language models to the multimodal space. LLaVA-RLHF~\citep{sun2023aligning}, the first open-source study in this area, improves visual chat abilities of LMMs using human-scored rankings. 
BPO~\citep{pi2024strengthening} constructs preference data by introducing negative responses generated by the model itself, using distorted images or text-based LLMs to inject errors. \citet{wang2024mdpoconditionalpreferenceoptimization} proposes mDPO, which uses conditional preference optimization to emphasize image information. Other works apply preference alignment to reduce hallucinations and enhance the overall capabilities of vision-language models (VLMs), either through human feedback (e.g., RLHF-V~\citep{yu2024rlhf}) or AI feedback (e.g., Silkie: VLFeedback~\citep{li2023silkie}).
Several approaches use self-rewarding mechanisms to minimize dependence on external preference pairs, such as divide-and-conquer strategies~\citep{yu2024rlaif} (RLAIF-V), sentence-level beam search~\citep{zhou2024calibrated}, deliberate hallucination injection~\citep{zhou2024aligning}, or metric-based self-critic prompts~\citep{wang2024enhancing}. As a generalist evaluator, LLaVA-Critic can provide valuable feedback for LMM alignment, paving the way for self-improving AI models.

\section{Data Collection}
\label{sec:data}

\begin{figure*}[!tp]
\centering
\begin{minipage}{.25\textwidth}
\includegraphics[height=120pt]{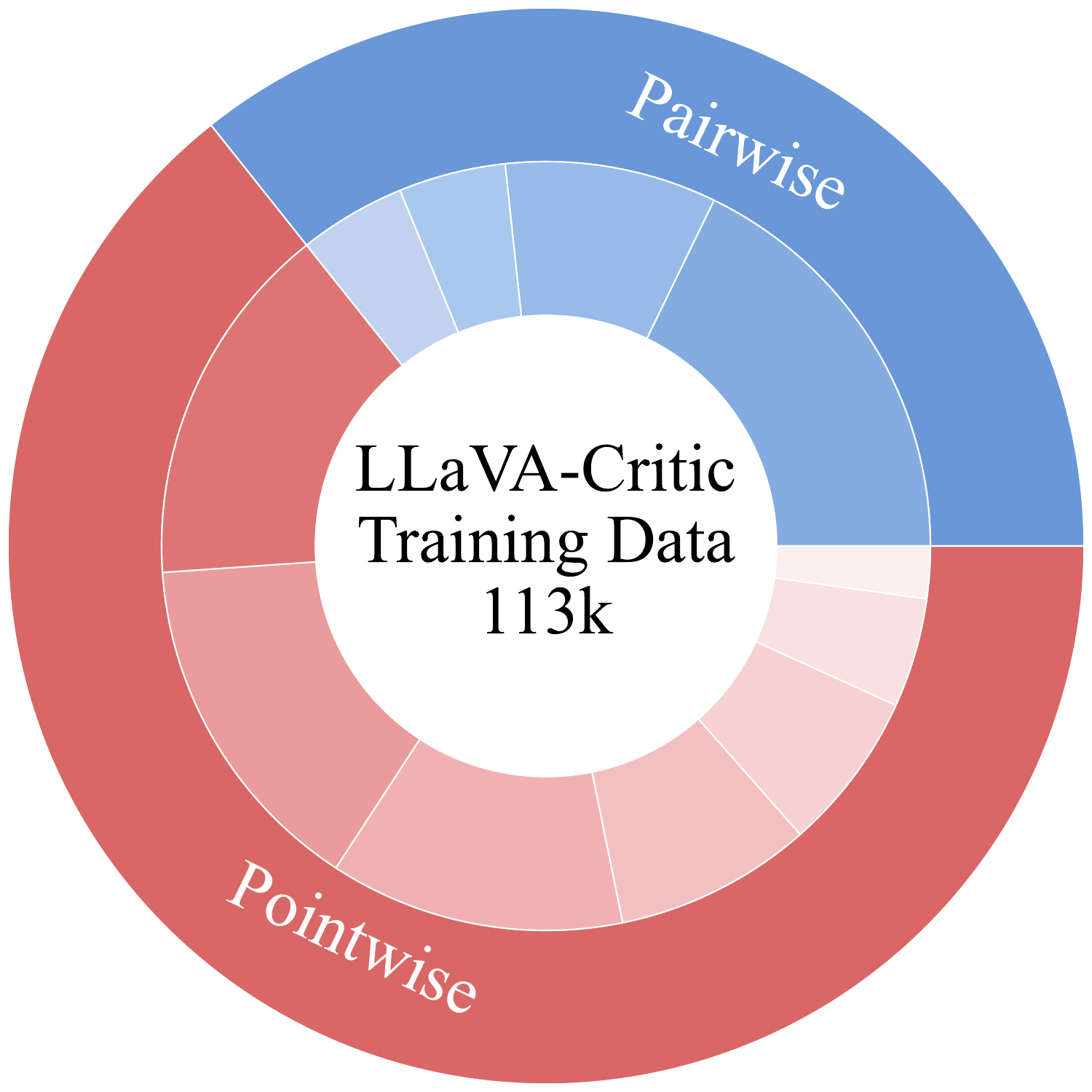}
\end{minipage}%
\hspace{1mm}
\begin{minipage}{.69\textwidth}
\centering
\renewcommand{\arraystretch}{1.18}
\resizebox{0.98\textwidth}{!}{ 
\setlength\tabcolsep{1pt}
\fontsize{6.8pt}{6pt}\selectfont
\begin{tabular}{c|l|lc|r}
\toprule 
\textbf{Setting} & \multicolumn{1}{c|}{\textbf{Prompt source}} & \multicolumn{2}{c|}{\textbf{Data source}} & \textbf{Data size}  \\  
\midrule
            &   LLaVA-in-the-Wild           & \tikz[baseline=0.05em] \fill [color={rgb,255: red,208; green,122; blue,120}, opacity=1] (0,0) rectangle (0.75em,0.75em); & LLaVA, SVIT, LLaVAR, LLaVAMed, ComVint& 17.5k   \\
            &   LLaVA-Wilder                & \tikz[baseline=0.05em] \fill [color={rgb,255: red,223; green,159; blue,157}, opacity=1] (0,0) rectangle (0.75em,0.75em); & SVIT, LLaVAR, LLaVAMed, ComVint, M3IT, PCAEval &  16.6k \\
            &   WildVision-Bench            & \tikz[baseline=0.05em] \fill [color={rgb,255: red,231; green,180; blue,180}, opacity=1] (0,0) rectangle (0.75em,0.75em);     & VLFeedback &  14.0k    \\
Pointwise   &   MMVet                       & \tikz[baseline=0.05em] \fill [color={rgb,255: red,235; green,194; blue,195}, opacity=1] (0,0) rectangle (0.75em,0.75em);     & LLaVAR, LLaVAMed, M3IT, PCAEval & 9.3k  \\
            &  MMHal-Bench                  & \tikz[baseline=0.05em] \fill [color={rgb,255: red,240; green,209; blue,211}, opacity=1] (0,0) rectangle (0.75em,0.75em); & LRV-Instruction & 7.6k  \\
            &   ImageDC                     & \tikz[baseline=0.05em] \fill [color={rgb,255: red,245; green,224; blue,225}, opacity=1] (0,0) rectangle (0.75em,0.75em); & SVIT-detail & 5.3k  \\
            &   RefoMB                      & \tikz[baseline=0.05em] \fill [color={rgb,255: red,251; green,239; blue,241}, opacity=1] (0,0) rectangle (0.75em,0.75em);     & VLFeedback &  2.5k   \\
            
\midrule
\multirow{4}{*}{Pairwise}  & \multicolumn{1}{c|}{\multirow{4}{*}{\makecell[c]{30 manually crafted \\prompt templates}}}    &\tikz[baseline=0.05em] \fill [color={rgb,255: red,140; green,171; blue,221}, opacity=1] (0,0) rectangle (0.75em,0.75em);  & VLFeedback              &  20.0k   \\
                            &                                                                       &\tikz[baseline=0.05em] \fill [color={rgb,255: red,157; green,184; blue,227}, opacity=1] (0,0) rectangle (0.75em,0.75em);    & LLaVA-RLHF           &  9.4k  \\
                            &                                                                       & \tikz[baseline=0.05em] \fill [color={rgb,255: red,174; green,199; blue,234}, opacity=1] (0,0) rectangle (0.75em,0.75em);  & VLFeedback (Tie)           &  5.0k   \\
                            &                                                                       &\tikz[baseline=0.05em] \fill [color={rgb,255: red,195; green,209; blue,235}, opacity=1] (0,0) rectangle (0.75em,0.75em);   &  RLHF-V           &  5.7k  \\
\bottomrule
\end{tabular}
}
\end{minipage}
\vspace{-2mm}
\caption{Data statistic of \ours-113k training dataset. In the pointwise setting, we categorize datasets by instruction sources and select data based on the task type corresponding to each evaluation prompt. Note that all our training data is sourced from public instruction-following training sets and does not overlap with with any evaluation benchmarks.}
\vspace{-5.5mm}
\label{fig:data_statistics}
\end{figure*}

We now introduce the data collection process for the \ours training dataset. 
The use of GPT-4/4V as a generalist evaluator for LMMs can be broadly categorized into two settings:  
$(i)$ {\bf Pointwise scoring}: GPT assigns a score to an individual candidate response, either by directly evaluating it based on predefined criteria or by scoring it relative to a fixed reference answer. This setting can be regarded as a combination of the {\it single-answer grading} and {\it reference grading} methods discussed in \citet{zheng2024judging}. 
$(ii)$
{\bf Pairwise ranking}: GPT compares two candidate responses to determine their relative quality (or declares a tie).
To equip \ours with a generalized evaluation capacity as with GPT-4V, we design a GPT-assisted pipeline to curate our training dataset for both settings. An example of \ours training data is shown in Table~\ref{fig:data_statistics}.

\definecolor{lightyellow}{rgb}{0.97,0.97,0.8}
\definecolor{lightpurple}{rgb}{0.99,0.92,0.97}
\definecolor{darkred}{rgb}{0.95, 0.4, 0.4}
\begin{table*}[t!]
\centering
\begin{minipage}{0.95\linewidth}\vspace{0mm}    
\centering
\begin{tcolorbox} 
\vspace{-1mm}
    \centering
    \footnotesize
    \begin{tabular}{p{0.97\columnwidth} c}
   \VarSty{ {\bf Multimodal Input for LMM:} } & \hspace{-3.2cm} \multirow{5}{*}{ \includegraphics[height=1.9cm]{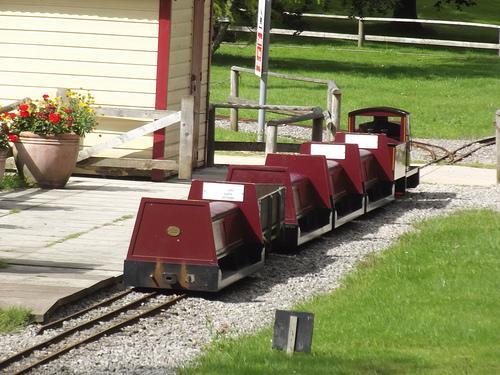} }\\
\textbf{\textit {Question}}: What are the specifics visible in the image? 
& 
\\
\rule{9cm}{0.4pt} & \\
\VarSty{ {\bf Setting 1: Pointwise Scoring} } & \\

\textbf{\textit {Response}}: 
The image shows a small train with four red cars, traveling on& \\ 
a track. The train is located in a park setting, and there are potted plants nearby. \\
\cellcolor{lightpurple}{\textbf{\textit{Evaluation Prompt}}: From 0 to 100, how much do you rate for this Text Caption in terms of the correct and comprehensive description of the image? Do not dominate the rating by a single attribute such as recognition correctness, but a overall rating on the object/scene appearance, position, pose, action, shape, etc., and contents in the background. Do not consider the appropriateness or sensitive descriptors, such as ``middle-aged western man'', judge based on if it has correct specifications of the object and scenes in image. Provide a few lines for explanation and the rate number at last after ``Final Score:''.} \\
\cellcolor{lightyellow}{\textbf{\VarSty{ {\bf LLaVA-Critic}}}:
The text caption provides a generally accurate description of the image. It correctly identifies the small train with four red cars and mentions the park setting and the presence of potted plants. However, there are a few details that could be improved for a more comprehensive description. For instance, the caption could mention that the train is stationary rather than traveling, and it could describe the surrounding structures, such as the building and the wooden fence, to give a fuller picture of the scene.} \textcolor{red}{Final Score: 85.} \\
\hrulefill & \\
   \VarSty{ {\bf Setting 2: Pairwise Ranking} } & \\
   \textbf{\textit {Response A}}: The image shows a small train on a track, with a total of six cars attached to it. The train is traveling on a small track, and there are potted plants nearby. The scene appears to be set in a park or a similar outdoor area.\\
   \textbf{\textit {Response B}}: The image shows a small train with four red cars, traveling on a track. The train is located in a park setting, and there are potted plants nearby.\\
   \cellcolor{lightpurple}{\textbf{\textit{Evaluation Prompt}}: ``Given an image and a corresponding question, please serve as an unbiased and fair judge to evaluate the quality of the answers provided by a Large Multimodal Model (LMM). Determine which answer is better and explain your reasoning with specific details. Your task is provided as follows:}\\
   \cellcolor{lightpurple}{Question: []\textbackslash n Response A: []\textbackslash n  Response B: [{}]\textbackslash n ASSISTANT:\textbackslash n''}\\
   \cellcolor{lightyellow}{\textbf{\VarSty{ {\bf LLaVA-Critic}}}: \textcolor{red}{Response B is better than response A} because it provides a more accurate description of the objects and their attributes in the image. Response B correctly identifies the train as having four red cars, whereas Response A inaccurately states that there are six cars. Both responses mention the potted plants and the park setting, but Response B's accuracy in the number of train cars makes it more precise. Additionally, Response B avoids unnecessary repetition and provides a clearer depiction of the scene, making it more helpful and accurate overall.} \\
    \end{tabular}
\end{tcolorbox}
\vspace{-4mm}
\caption{An example of \ours training data. The top block shows \textcolor{red}{pointwise scoring}, where \ours predicts a score to evaluate a single response's quality; the bottom block illustrates \textcolor{red}{pairwise ranking}, where it rank response pairs. In both settings, \ours learns to provide reasons for its judgments.}
\vspace{-5mm}
    \label{tab:full_example_car_bbox}
\end{minipage}
\end{table*}

\subsection{Pointwise Data}
\label{sec: pointwise_data}

To train a generalist critic model for the evaluation of individual LMM responses, each pointwise training data instance consists of the following components:  a multimodal instruction (i.e., a question-image pair), a model response, an optional reference answer (depending on the evaluation prompt), an evaluation prompt, a judgment score, and the corresponding justification for the score. By organizing them in a sequence, the training sample is:

\vspace{1mm}
\noindent\texttt{(Image, Question, Response, Reference,}\\
\texttt{Evaluation Criteria, {\color{mygreen} Score}, {\color{mygreen} Reason})}
\vspace{1mm}

\noindent where green parts are treated as model output to compute the auto-regressive loss, the order of $\texttt{Score}$ and $\texttt{Reason}$ is specified by the evaluation prompt.
We select multimodal instructions from 8 multimodal instruction tuning datasets, spanning across a wide range of tasks including: 
(1) general visual conversation, detailed captioning and reasoning (LLaVA-Instruction-150k~\citep{liu2023llava}, SVIT~\citep{zhao2023svit});
(2) more challenging tasks such as complex reasoning (ComVint~\citep{du2023makes}), text-rich understanding (LLaVAR~\citep{zhang2023llavar}) and robustness-oriented instructions (LRV-Instruction~\citep{liu2023aligning});
and (3) various specific domains such as academic question answering (M3IT~\citep{li2023m}), medical image understanding (LLaVA-Med~\citep{li2023llavamed}) and embodied decision-making (PCA-EVAL~\citep{chen2023towards}). 
For each multimodal instruction, we select one or more model responses from VLFeedback~\citep{li2023silkie}, which collects multiple responses from 12 off-the-shelf LMMs. Additionally, we generate responses using GPT-4o, a leading commercial LMM, to serve as high-quality reference answers.

To equip \ours with general evaluation capacities across various tasks, we construct an evaluation prompt pool from 7 widely used multimodal benchmarks that utilize GPT-as-a-judge, including LLaVA-in-the-Wild~\citep{liu2023llava}, LLaVA-Wilder~\citep{li2024llavanext-strong}, Image Detailed Captioning~\citep{li2024llavanext-strong}, MMHal-Bench~\citep{sun2023aligning}, MMVet~\citep{yu2023mmvetevaluatinglargemultimodal}, WildVision-Bench~\citep{lu2024wildvision} and RefoMB~\citep{yu2024rlaif}.
\footnote{
Although RefoMB and WildVision-Bench use pairwise evaluation prompts, only one response is evaluated, with the other from a fixed reference model (GPT-4V and Claude-3-Sonnet, respectively), making them pointwise evaluations. 
} 
Prompts that require additional textual context—since they use text-only GPT-4 as the evaluator—are adjusted to focus on the input image, better aligning with the LMM evaluator setting. To construct training data based on each evaluation prompt, we select multimodal instructions and model responses  according to the specified evaluation scenario, and include reference answers from GPT-4o when necessary. These components are then assembled into the evaluation prompt and used as input for GPT-4o (as-a-judge) to provide high-quality judgment scores and detailed justifications for model responses. Finally, our pointwise training dataset comprises a total of 18,915 question-image pairs and 72,782 critic data samples.

\subsection{Pairwise Data}\label{sec: pairwise_data}
The pairwise data consists of responses with known preference relationships. We collect pairwise data from three datasets: VLFeedback~\citep{li2023silkie}, RLHF~\citep{sun2023aligning}, and RLHF-V~\citep{yu2024rlhf}. 
In the VLFeedback dataset, each (question, response) pair is rated across three different dimensions by GPT-4V.
For the same question, responses generated by different LMMs can form multiple response pairs for that question. 
We randomly select 20k pairs where the average score gap between responses is greater than 0.6. 
Besides, to ensure diversity in the preferences, we randomly sample 5k pairs where the two responses had identical scores across all three dimensions to serve as ``Tie'' training data.
In the RLHF dataset, each question is annotated with preference relationships between different responses by human evaluators. In contrast, the RLHF-V dataset consists of responses generated by LMM, which have been manually refined to produce improved responses. From these two datasets, we collect 9.4k (RLHF) and 5.7k (RLHF-V) response pairs, each annotated with human preferences. This results in a total of 40.1k pairwise data samples.

To enable \ours to provide useful detailed feedback in addition to the preference relation, we utilize GPT-4o to generate reasons behind the given preference judgment. 
The training sample for pairwise data is structured in the following sequence: 

\vspace{1mm}
\noindent\texttt{\small(Image, Question, Response 1\&2, Evaluation)} \\ 
\texttt{\small(Criteria, {\color{mygreen} Preference}, {\color{mygreen} Reason})} 
\vspace{1mm}

\noindent where the evaluation criteria is from carefully designed prompt templates.
To allow \ours to handle diverse pairwise ranking, we develop a set of 30 evaluation prompt templates (see Appendix~\ref{append:training_prompts}). Each preference pair is randomly assigned a template from this set to form the final training data.

\vspace{-4mm}
\paragraph{Data statistics.}

Our training dataset comprises a total of 46k images and 113k data samples. As shown in Figure~\ref{fig:data_statistics}, we curate our
training set with diverse instruction-response pairs, spanning multiple evaluation tasks and domains.

\vspace{-0.5mm}
\section{LLaVA-Critic}
\label{sec:llava_critic}

\subsection{Model}
\label{sec:model}
To train the \ours model, we fine-tune a pre-trained LMM that already possesses strong capabilities in following diverse instructions. This is crucial, as it ensures that the model has already been equipped to handle a wide range of vision tasks in the wild with high quality. The evaluation ability is treated as an additional discriminative ability closely tied to these scenarios. 
During training, \ours takes an evaluation prompt---assembling the multimodal instruction input, model response(s), and an optional reference response---as input. It is trained to predict quantitative pointwise scores or pairwise rankings based on the criteria in the evaluation prompt, and provide detailed justifications for the assigned judgments. Standard cross-entropy loss is applied to both judgments and justifications. 

In experiment, we start with the LLaVA-OneVision(OV) 7B/72B pretrained checkpoint and fine-tune it on the proposed \ours-113k dataset for 1 epoch to develop \ours. We apply a learning rate of 2e-6 and a batch size of 32 for training, with other hyperparameters set to the defaults from \citet{li2024llavaonevision}. 
We also curate a subset with 53k samples (42k pointwise, 11k pairwise) that cover fewer instruction sources and domains. The model trained on this reduced subset is referred to as \ours (v0.5). 

%

\subsection{Scenario 1: LMM-as-a-Judge}
\label{sec:method-lmm-judge}
Evaluating complex tasks often requires human judges to provide feedback, which can be labor-intensive. LLaVA-Critic can serve as a general evaluator for LMM responses, reducing labor costs by automating the evaluation process.
\ours consistently provides reliable judgments and justifications aligned with GPT-4o or human evaluations across a range of widely used multimodal benchmarks. This consistency holds true for both instance-level scoring and model-level ranking, as demonstrated in Sec.~\ref{exp:lmm-judge}.

Specifically, we consider the following evaluation scenarios: 
$(i)$ \textit{Visual Chat}. This task involves handling daily-life visual tasks through multimodal dialogue, requiring evaluation of task completion quality in a conversation setting. Examples include LLaVA-Bench~\citep{liu2023llava} and LLaVA-in-the-Wild~\citep{liu2023llava}, which focus on simpler scenarios, while LLaVA-Wilder~\citep{li2024llavanext-strong} addresses more challenging cases. 
$(ii)$ \textit{Integrated capabilities}. Real-world tasks require integration of multiple basic abilities of LMMs.
MM-Vet~\citep{yu2023mmvetevaluatinglargemultimodal} offers a comprehensive benchmark, evaluating core vision-language capabilities including recognition, OCR, knowledge integration, language generation, spatial awareness, and math. The Multimodal Live-Bench tests the model's ability to generalize to new, unobserved knowledge by leveraging continuously updated news and online forums.
$(iii)$ \textit{Preferences}. This task simulates real-world multimodal interactions where models are expected to align their behavior with human preferences. The WildVision-Bench~\citep{lu2024wildvision} is a prime example, replicating scenarios from the online platform WildVision-Arena (WV-Arena) to evaluate preference-based interactions.
$(iv)$ \textit{Detailed Description}.  
This task assesses models on their ability to provide comprehensive and detailed descriptions of images and videos. Image Detailed Captioning~\citep{li2024llavanext-strong} evaluates detailed descriptions in images, while Video Detailed Captioning~\citep{zhang2024llavanext-video} extends these capabilities to video contexts.
$(v)$ \textit{Hallucination}: This task focuses on the model's ability to provide grounded responses based on the given context, ensuring that it avoids generating inaccurate or fabricated information, exemplified by MMHal-Bench~\citep{sun2023aligning}.

\subsection{Scenario 2: Preference Learning}
\label{sec:method-dpo}

Leveraging a generalist evaluator as a critic to generate reward signals for reinforcement learning is a promising research direction. In this work, we employ LLaVA-Critic to produce AI-generated feedback datasets for diverse tasks, thereby improving the performance of supervised fine-tuned LMMs through preference alignment. Notably, the reward signals generated by our critic can be utilized in any preference learning algorithms, including RLHF and DPO. To quickly assess the effectiveness of the reward data, we focus on how \ours is incorporated into the iterative DPO training process.

\begin{itemize}[leftmargin=2.5mm]
\setlength{\itemsep}{2pt}
\item \textit{\textbf{Step 1: Response generation}.} The iterative DPO process begins with a pretrained LMM $\pi_0$ as the initial checkpoint and a set of multimodal instructions $\{(x_k, v_k)\}_{k=1}^N$, where each $x_k$ is a question and $v_k$ is the corresponding image.
For each question-image pair $(x, v)$, the pretrained LMM $\pi_0$ randomly generates $K$ responses $\{y_1, y_2, \dots, y_K\}$, sampled independently from its distribution.

\item \textit{\textbf{Step 2: Scoring}}.
To mitigate order-related variance in \ours's preferences, we form all possible ordered pairs from these responses, resulting in $K \times (K - 1)$ pairs. For each response pair $(y_i, y_j)$, we apply \ours with an evaluation prompt to generate a relative score $a_{ij}$, which normalizes the score of $y_j$ based on $y_i$.

\item  \textit{\textbf{Step 3: Reward Preference}.} 
The overall reward score $r_i$ for each response $y_i$ is calculated by aggregating these preference scores:
$
    r_i = \sum_{k\neq i} a_{ki} - \sum_{l\neq i} a_{il}
$
This calculation effectively measures how much better or worse $y_i$ is compared to all other responses.
We then select the responses with the highest and lowest reward scores as the best and worst responses, denoted as $y^{+}$ and $y^{-}$, respectively. These form the pairwise feedback data $(y^{+}, y^{-})$, which is used for DPO training to enhance the LMM's alignment with \ours's preferences.

\end{itemize}

\vspace{-4mm}
\paragraph{Iterative Improvement.}  After each round of DPO training, the updated LMM becomes the new starting checkpoint. The process is then iterated for another $M-1$ rounds, using LLaVA-Critic to progressively improve the model's performance based on its self-generated responses.
\section{Experimental Results}
\label{sec:experimental}

\subsection{LMM-as-a-Judge}
\label{exp:lmm-judge}

To comprehensively assess \ours's capacity in evaluating LMM responses across different scenarios, we consider two primary experimental settings:
(1) {\it In-domain Judgments}: where we measure \ours's consistency with GPT-4o or human evaluators on evaluation tasks/prompts included in the \ours-113k dataset; 
and (2) {\it Out-of-domain Judgments}: where we apply \ours on tasks and prompts that are unseen during training. For the second setting, we use the MLLM-as-a-Judge~\citep{chen2024mllm} benchmark to assess the alignment between LLaVA-Critic and human evaluators in generalized scenarios.

\vspace{-2mm}
\paragraph{In-domain Pointwise Scoring}
To evaluate the consistency between \ours and GPT-4o~\citep{openai2024gpt4o} in pointwise scoring across different evaluation scenarios, as described in Sec.~\ref{sec:method-lmm-judge}, we select 7 popular multimodal benchmarks and collect candidate responses from 13 commonly used LMMs alongside their GPT-4o evaluations, resulting in a  total of 14174 examples (see details in Appendix~\ref{append:indomain-pointwise-eval}). \ours is then tasked with providing judgments on theses samples. We report Pearson correlation to measure the degree of alignment with GPT-4o in instance-level scoring.

We conduct experiments based on three different baseline models: LLaVA-NeXT (LLaMA-8B)~\citep{liu2024llavanext, li2024llavanext-strong}, LLaVA-OneVision-7B, and LLaVA-OneVision-72B. As shown in Table~\ref{tab:result_in_domain_pointwise},
\ours variants significantly outperform their corresponding baseline models across all models and benchmarks. 
$(i)$ \textit{Data scaling}.
By comparing the performance between v0.5 and full data trained LLaVA-Critic-7B, it concludes the necessity of larger size and diversity of instruction in training data.
$(ii)$ \textit{Model scaling}.
The best performance in terms of Pearson-r is achieved by LLaVA-Critic-72B with an average score of 0.754, which significantly outperforms the LLaVA-OV-72B baseline (0.634). 
This indicates that LLaVA-Critic-72B already possesses pointwise scoring capabilities that are quite aligned with GPT-4o.
Despite a considerable reduction in model size, LLaVA-Critic-7B retains very strong point-wise scoring capabilities. 
With a score of 0.732, its shows minimal performance decline compared to \ours-72B, and significantly outperforms other advanced open-source LMMs of similar size, such as Qwen2-VL (0.352) and LLaMA3.2-Vision (0.359).
This presents an advantage for deploying and utilizing LLaVA-Critic in resource-constrained environments. 
We also provide Kendall's Tau results in Appendix~\ref{append:pointwise-kendall-tau} to assess model-level ranking, which reveal similar patterns and conclusions.

\definecolor{front-color}{HTML}{F5FFFA}
\begin{table*}[!tp]
\vspace{-5mm}
\centering
\setlength{\tabcolsep}{1.5pt}
\small
\scalebox{0.97}{
\begin{tabular}{l|ccccccc|c}
\toprule
\multirow{2}{*}{LMM Evaluator} & \multicolumn{8}{c}{Pearson-r ($\uparrow$)} \\
\cmidrule{2-9}
 & ImageDC & MMVet & WildVision & LLaVA-B & LLaVA-W & L-Wilder & MMHal & Avg. \\
\midrule 
LLaVA-NeXT (LLaMA-8B) & 0.262 & 0.317 & 0.147 & 0.211 & 0.345 & 0.156 & 0.472 & 0.273  \\
\rowcolor{front-color}
LLaVA-Critic (LLaVA-NeXT) & 0.673 &	0.706 &	0.580	& 0.529 & 	0.820 &	0.936 & 0.748& 0.713  \\ \hline
Qwen2-VL-7B-Instruct & 0.199 & 0.463 & 0.096 & 0.208 & 0.476 & 0.694 & 0.329 & 0.352 \\
LLaMA3.2-11B-Vision-Instruct & 0.069 & 0.450 & 0.224 & 0.356 & 0.499 & 0.531 & 0.387 & 0.359 \\
LLaVA-OV-7B  & 0.056 & 0.349 & 0.251 & 0.335 & 0.533 & 0.592 & 0.433 & 0.364 \\
\rowcolor{front-color}
LLaVA-Critic-7B (v0.5)  & 0.737  & 0.718  & 0.571  & 0.494  & 0.789  & 0.932  & 0.746  & 0.712 \\
\rowcolor{front-color}
LLaVA-Critic-7B  & 0.735 & 0.733 & 0.616 & 0.510 & 0.843 & 0.940 & 0.748 & 0.732 \\ 
\hline
LLaVA-OV-72B & 0.718 & 0.680 & 0.446 & 0.436 & 0.716 & 0.824 & 0.620 & 0.634 \\
\rowcolor{front-color}
LLaVA-Critic-72B & 0.802 & 0.723 & 0.705 & 0.524 & 0.782 & 0.951 & 0.790 & 0.754 \\
\bottomrule
\end{tabular} }\\
\vspace{-2mm}
\caption{Results on in-domain pointwise scoring. \ours consistently outperforms baselines across 7 multimodal benchmarks.}
\label{tab:result_in_domain_pointwise}
\end{table*}

\begin{figure*}[!t]
    \centering
    \vspace{-2mm}
    \includegraphics[width=0.9\linewidth]{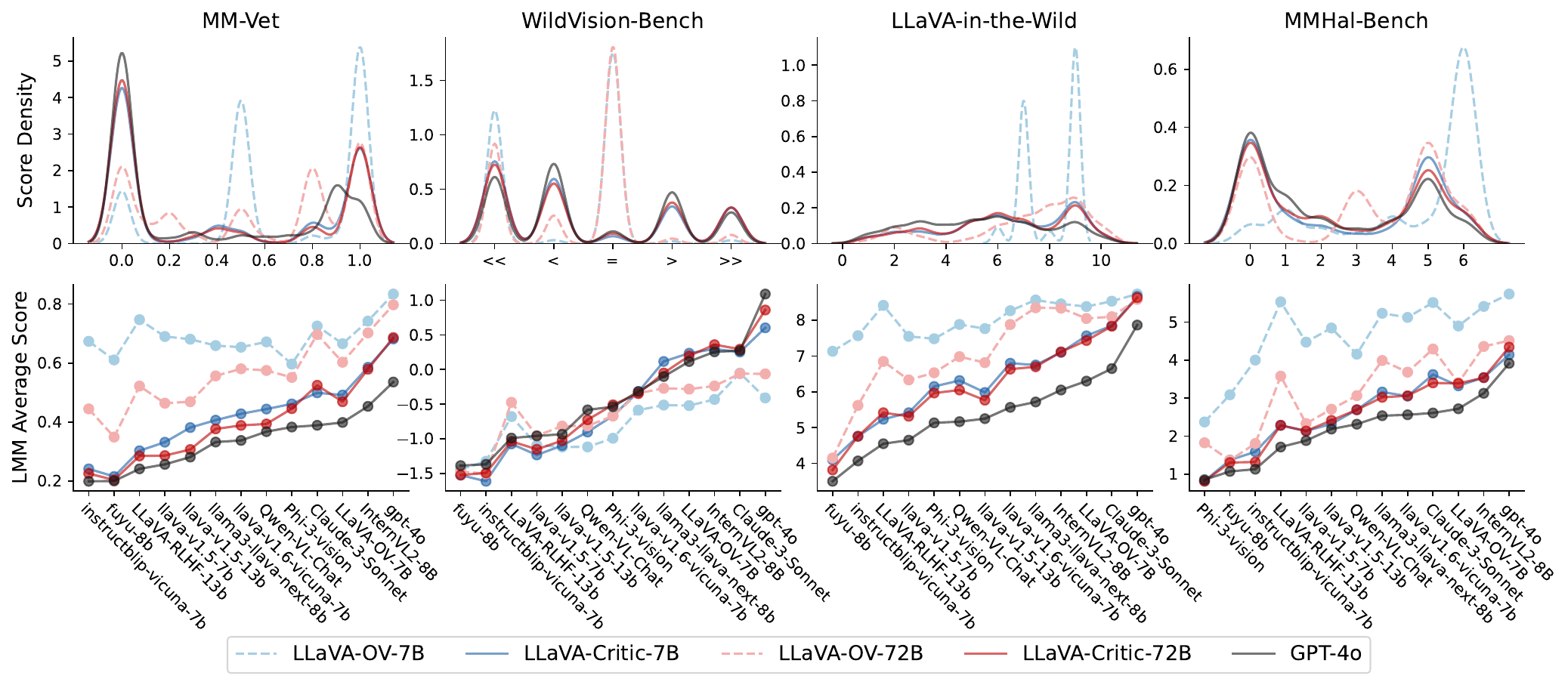}
    \vspace{-3mm}
    \caption{\textit{(Top):} Overall distribution of evaluation scores across 4 benchmarks. \textit{(Bottom):} Calculated average evaluation score for each response model on each benchmark. Each color represents a different LMM evaluator. Leveraging high-quality critic training data, \ours closely aligns with GPT-4o in delivering balanced evaluation scores and accurately ranking response LMMs. } 
    \label{fig:pointwise_score_fig}
    \vspace{-3mm}
\end{figure*}

Figure~\ref{fig:pointwise_score_fig} presents a qualitative comparison between \ours and other LMM evaluators. While LLaVA-OneVision often assigns fixed scores (e.g., ``Tie'' on WildVision-Bench or ``6'' on MMHal-Bench), \ours produces more diverse and balanced scores that closely align with GPT-4o, leading to consistent rankings of response models. 
Notably, even without training on critic data, LLaVA-OneVision-72B demonstrates model-level rankings that partially align with GPT-4o across four multimodal benchmarks. 

\vspace{-4mm}
\paragraph{In-domain Pairwise Ranking} To assess the consistency between \ours and human evaluators in pairwise ranking, we use the battle data from WildVision Arena~\citep{lu2024wildvision}, which comprises 11k human-annotated preference relations among LMM response pairs.
Each relation includes a question-image pair and two responses generated by different models, accompanied by a human-annotated preference (including ties). 
From this dataset, we randomly sample 2k response pairs and assign them to evaluation prompts from the pairwise ranking prompt template set mentioned in Section~\ref{sec: pairwise_data}, creating the in-domain evaluation dataset. 
We report average accuracy, with and without ties, to assess alignment with human evaluators at the instance level.
For model-level consistency, we calculate the Elo rating for each response LMM and report Kendall's Tau to measure the overall ranking correlation with human preferences.

Experimental results are reported in Table~\ref{tab:results_in_domain_pairwise}. While existing open-source LMMs exhibit initial pairwise ranking ability, there is a notable performance gap compared to GPT-4V/4o. After training with critic data, \ours achieves significant improvements. Specifically, \ours-72B achieves an average accuracy of 73.6\% in pairwise comparisons without tie, outperforming both GPT-4o and GPT-4V. For pairwise accuracy with tie and model-level ranking (Kendall's Tau), \ours-72B shows only a marginal gap compared to GPT-4V/4o, with an accuracy of 60.5\% and a score of 0.779, respectively. Notably, despite a substantial reduction in the number of parameters, \ours-7B still achieves an average accuracy of 59.6\% in pairwise ranking with ties and 72.2\%  without ties, alongside a Kendall's tau of 0.763. These results underscore the strong alignment between \ours and human evaluators in pairwise ranking LMM responses.

\vspace{-4mm}
\paragraph{MLLM-as-a-Judge} 
MLLM-as-a-Judge~\citep{chen2024mllm} is a comprehensive benchmark to evaluate the degree of alignment between model-based evaluation and human evaluation.
It collects approximately 17k image-instruction-response triplets across 14 multimodal benchmarks and 6 LMM response models. Human annotators are then employed to assess model responses under scoring, pairwise comparison and batch ranking settings, resulting in 7756, 5719, 1469 examples respectively. In our experiments, we evaluate \ours in both (pointwise) scoring and pair comparison settings to assess its general alignment with human evaluators. We report the average Pearson correlation for scoring and average accuracy for pairwise comparison, following the metrics used in the original benchmark. 

We compare \ours with commercial models (GPT-4V/4o, Gemini-Pro~\citep{team2023gemini}), open-source LMMs, as well as Prometheus-Vision~\citep{lee2024prometheusvision}, which trains a LLaVA model on a curated LMM-as-a-judge dataset comprising 15k GPT-generated rubrics and 150k GPT-4V feedback data. 
As demonstrated in Table~\ref{tab:mllm-as-a-judge}, \ours-7B surpasses all baselines of comparable model size by a substantial margin. Built on a stronger base model, \ours-72B further achieves the Pearson similarity with human annotators from 0.314 to 0.393 in pointwise scoring. For pairwise comparisons, it achieves accuracy rates of 57.8\% and 71.5\% with and without ties, respectively, reaching a level of alignment with human evaluators comparable to GPT-4V/4o. We also compare different variants of \ours and observe performance gains with both stronger base models and larger training data, consistent with previous findings. This again highlights the critical role of model and data scaling in building an effective and generalist open-source LMM evaluator. More comprehensive results are provided in Appendix~\ref{apppend:mllm-as-a-judge-results}.

\definecolor{light-gray}{gray}{0.9}
\begin{table}[!t]
\centering
\small
\setlength{\tabcolsep}{1.5pt}
\scalebox{0.94}{
\begin{tabular}{l|ccc}
\toprule
Model                      & Acc{\scriptsize(w. Tie)}$\uparrow$ & Acc{\scriptsize(w.o. Tie)}$\uparrow$ & Kendall's $\tau\uparrow$ \\
\hline
\rowcolor{light-gray}
GPT-4o                     & 0.617          & 0.734            &    0.819     \\
\rowcolor{light-gray}
GPT-4V                     & 0.620          & 0.733            &     0.787        \\
\hline
LLaVA-NeXT {\scriptsize(LLaMA-8B)}            & 0.473          & 0.569            &   0.605          \\
LLaVA-OV-7B               & 0.531          & 0.640            &     0.715        \\
Qwen2-VL-7B-Instruct & 0.550 & 0.678 & 0.699 \\
LLaMA3.2-V {\scriptsize(11B-Instruct)} & 0.513 & 0.673 & 0.737 \\
LLaVA-OV-72B               & 0.594          & 0.708            &    0.763         \\
\hline
\rowcolor{front-color}
LLaVA-Critic-7B {\scriptsize(v0.5)}     & 0.580          & 0.692            &     0.755      \\
\rowcolor{front-color}
LLaVA-Critic {\scriptsize(LLaVA-NeXT)}   & 0.582          & 0.686            &    0.755         \\
\rowcolor{front-color}
LLaVA-Critic-7B            & 0.596          & 0.722            &    0.763        \\
\rowcolor{front-color}
LLaVA-Critic-72B           & 0.605          & 0.736            &   0.779     \\
\bottomrule
\end{tabular}
}
\vspace{-1.5mm}
\caption{Results on in-domain pairwise ranking. \ours is comparable with GPT-4V in alignment with human evaluators.}
\label{tab:results_in_domain_pairwise}
\vspace{-6mm}
\end{table}

\vspace{-4mm}
\paragraph{Qualitative Comparison} We present example comparisons of the evaluation scores and reasons generated by \ours and other LMMs, with detailed examples provided in Appendix~\ref{append:more_examples}. The key findings are as follows: Compared to LLaVA-OneVision, \ours delivers more accurate judgments (Table~\ref{tab:critic_example_seven}), and provides more concrete, image-grounded justifications (Table~\ref{tab:critic_example_beach}). The latter is crucial for reliable AI~\citep{bai2022training}, as offering well-supported reasons for evaluations establishes \ours as a transparent evaluator of LMM responses.

\begin{table}[!ht]
\centering
\small
\setlength{\tabcolsep}{2pt}
\scalebox{0.96}{
\begin{tabular}{l|ccc}
\toprule
Model & Score$\uparrow$ & Pair{\footnotesize (w. Tie)}$\uparrow$ & Pair {\footnotesize (w.o. Tie)}$\uparrow$ \\
\midrule
\rowcolor{light-gray}
GPT-4V*  & 0.490 & 0.636 & 0.773 \\
\rowcolor{light-gray}
GPT-4o\textsuperscript{\textdagger} & 0.439 & 0.577 & 0.736 \\
\rowcolor{light-gray}
GPT-4V\textsuperscript{\textdagger} & 0.424 & 0.538 & 0.717 \\
\rowcolor{light-gray}
Gemini-pro* & 0.304 & 0.509 & 0.615 \\
\hline
LLaVA-v1.5-7B & 0.158 & 0.439 & 0.576 \\
Prometheus-V {\scriptsize(LLaVA-v1.5-7B)} & 0.213 & -- & -- \\
LLaVA-NeXT {\scriptsize(LLaMA-8B)} & 0.198 & 0.461 & 0.586 \\
LlaVA-OV-7B & 0.151 & 0.426 & 0.550\\
Qwen2-VL-7B-Instruct & 0.253 & 0.348 & 0.645 \\
LLaMA3.2-V {\scriptsize(11B-Instruct)} & 0.237 & 0.529 & 0.658 \\
LlaVA-OV-72B & 0.287 & 0.513 & 0.701\\

\hline
\rowcolor{front-color}
 LLaVA-Critic {\scriptsize(LLaVA-v1.5-7B)} & 0.228 & 0.528 & 0.656 \\
 \rowcolor{front-color}
 LLaVA-Critic {\scriptsize(LLaVA-NeXT)} & 0.272 & 0.547 & 0.677\\
 \rowcolor{front-color}
 LLaVA-Critic-7B {\scriptsize(v0.5)} & 0.312 & 0.546 & 0.675\\
 \rowcolor{front-color}
 LLaVA-Critic-7B  & 0.314 & 0.556 & 0.689\\
 \rowcolor{front-color}
 LLaVA-Critic-72B  & 0.393 & 0.578 & 0.715\\
\bottomrule
\end{tabular}
}
\vspace{-1.5mm}
\caption{Results on MLLM-as-a-Judge~\citep{chen2024mllm}.  *: the results as reported in the original paper~\citep{chen2024mllm}; \textsuperscript{\textdagger}: results from our evaluation of GPT-4V/4o based on their codebase. Note that Prometheus-Vision cannot follow the pairwise evaluation prompt. \ours significantly narrows the gap between open-source LMMs and GPT-4V/4o in their ability to evaluate LMM responses across a wide range of evaluation scenarios.}
\vspace{-5mm}
\label{tab:mllm-as-a-judge}
\end{table}

\vspace{-4mm}
\paragraph{Critic training preserves original visual capacities.} As shown in Appendix~\ref{append:critic_visual}, \ours largely preserves LLaVA-OV's original ability in handling diverse visual tasks and achieves modest gains in visual chat performance.

\subsection{Preference Learning}\label{sec: exp_pl}

\begin{table*}[!t]
    \centering
    \small
    \setlength{\tabcolsep}{2pt}
    \scalebox{0.98}{
    \begin{tabular}{l|c|ccccc|ccccc}
    \toprule
      Method   & \#Prompts & LLaVA-W & L-Wilder & WildVision & LiveBench & MMHal* &  $\text{MME}^{P}$ & $\text{MME}^{C}$ & MMB-en & MM-Vet & MMStar \\
    \midrule
      LLaVA-v1.5-7B  & -- & 63.4 & 54.2 & \underline{20.4} & 45.6 & 1.94 & \underline{1510.7} & 348.2 & 64.3 & 31.1 & 33.3 \\
      + RLHF & 9.4k & 63.7 & 54.5 & 19.8 & 46.2 &  1.90 & 1508.2 & 360.2 & 60.4 & 31.1 & 33.0\\
      + SIMA & 17k &  66.1 & 52.3 & 17.6 & 47.9 &  1.81 & 1507.7 & \textbf{379.3} & \underline{64.9} & 31.6 & \underline{34.7}\\
      + CSR & 15k & 71.1 & 55.9 & 20.0 & 45.0 & 1.96 & \textbf{1524.2} & \underline{367.9} & \textbf{65.4} & \textbf{33.9} & 33.6\\
      + RLAIF-V & 33.8k & \underline{72.7} & \underline{56.4} & 19.2 & \textbf{50.4} &  \textbf{3.04} & 1362.7 & 302.9 & 62.6 & 26.7 & \textbf{35.4} \\
      \rowcolor{front-color} 
      + \ours & 9.4k & \textbf{73.5} & \textbf{57.2} & \textbf{29.2} & \underline{50.0} & \underline{2.07} & 1500.4 & 350.7 & 64.1 & \underline{32.2} & 34.2  \\
    \bottomrule
    \end{tabular}
    }
    \vspace{-1.5mm}
    \caption{Comparison with other preference learning algorithms on LLaVA-v1.5-7B. Apart from benchmarks in Table~\ref{tab:results_iter_dpo}, we also report the results on 4 comprehensive multimodal benchmarks for reference. The best and second best results are shown in \textbf{bold} and \underline{underlined} respectively. *OpenAI's \textit{gpt-4-0613} is used for the MMhal-Bench evaluation due to the deprecation of the original API.}
    \vspace{-1mm}
    \label{tab:dpo_llava15}
\end{table*}

We further evaluate \ours's performance in providing reward signals for iterative DPO.
LLaVA-OneVision's supervised fine-tuned checkpoint is used as the base policy model, and question-image pairs from LLaVA-RLHF~\citep{sun2023aligning} serve as the multimodal instructions.
For each pair, $K=5$ candidate responses are generated through random decoding (with a temperature of 0.7 and top-p of 0.9) to ensure response diversity. \ours is employed as described in  Sec. \ref{sec:method-dpo} to construct the pairwise feedback data, which is then used for one epoch of DPO training. We perform iterative DPO for $M=3$ rounds in total. 

To assess the effectiveness of \ours's reward signals, we evaluate the final LMM checkpoint on 6 open-ended multimodal benchmarks: four image-based tasks (LLaVA-in-the-Wild~\citep{liu2023llava}, LLaVA-Wilder~\citep{li2024llavanext-strong}, WildVision-Bench~\citep{lu2024wildvision} and LiveBench~\citep{zhang2024lmmseval}), one video-based task (Video Detailed Captioning~\citep{li2024llavanext-strong}), and one hallucination benchmark (MMHal-Bench~\citep{sun2023aligning}). We compare \ours with two baselines: (1) reward model from LLaVA-RLHF~\citep{sun2023aligning},  which is trained on human preferences, and (2) a naive baseline that replaces LLaVA-Critic with LLaVA-OneVision as a zero-shot reward model.

As shown in Table~\ref{tab:results_iter_dpo}, preferences provided by \ours significantly improve LLaVA-OneVision's visual chat capacities and reduce hallucination across challenging tasks.
\ours consistently surpasses other baseline reward models on 5 out of 6 benchmarks for the 7B base model and all 6 benchmarks for the 72B base model. Despite the preference alignment conducted solely with images, \ours also enhances LLaVA-OneVision's performance in Video Detailed Captioning (+0.12 on OV-7B and +0.26 on OV-7B), demonstrating its ability to generalize to both image and video contexts. Additionally, we observe that Critic-7B outperforms Critic-7B-v0.5 on 5 out of 6 benchmarks, highlighting the importance of stronger reward models---trained on more diverse critic instructions---to deliver more accurate reward signals and further enhance preference learning. (See Appendix~\ref{append:dpo_results} for additional results and Table~\ref{tab:example_chat_dpo} for a visual-chat example.) Notably, while using OpenAI’s GPT-4o as a reward model for 3 rounds of iterative DPO would cost approximately \$690, \ours provides a reliable, cost-free alternative for providing reward signals.

\begin{table}[!t]
\setlength{\tabcolsep}{1pt}
\centering
\small
\vspace{-3mm}
\scalebox{0.85}{
\begin{tabular}{l|l|cccccc}
\toprule
 Base & Reward  &  LLaVA-W & L-Wilder & WV-B & Live-B & V-DC & MMHal \\
\midrule
\rowcolor{light-gray}
GPT-4V & -- & 98.0 & 81.0 & 79.8 & 73.7 & 4.00 & 3.83 \\
\midrule
\multirow{5}{*}{OV-7B} & -- & 90.7 & 67.8 & 54.0 & 77.1 &  3.75 & 3.19 \\
 & OV-7B & 98.6 & 70.9 &  66.6 & 84.0 & 3.77 & 3.79\\
 & LLaVA-RLHF & 97.5 & 70.3 & 64.1  & 83.1 &  3.84 & \textbf{4.01} \\
\rowcolor{front-color}
\cellcolor{white} & Critic-7B {\scriptsize (v0.5)}  & 98.1& 70.5& 67.2 & \textbf{85.1}  &  3.83 & 3.85 \\
 \rowcolor{front-color}
\cellcolor{white} & Critic-7B  & \textbf{100.3} & \textbf{71.6} & \textbf{67.3} & 84.5 & \textbf{3.87} & 3.91 \\
\hline
\multirow{3}{*}{OV-72B} & --  & 93.5 & 72.0 & 51.7 & 81.5 & 3.60 & 3.61 \\
&  LLaVA-RLHF & 103.2 & 75.2 & 65.2 & 86.2 & 3.85 & 3.67\\
\rowcolor{front-color}
\cellcolor{white} &  Critic-72B   & \textbf{104.4} & \textbf{75.9} & \textbf{70.0} & \textbf{88.5} & \textbf{3.86} & \textbf{3.77}\\

\bottomrule
\end{tabular}
}
\vspace{-1.5mm}
\caption{Comparison between LLaVA-Critic and baselines in preference alignment. ``Base'': the initial LMM checkpoint for DPO.  
}
\vspace{-5mm}
\label{tab:results_iter_dpo}
\end{table}

\vspace{-3mm}
\paragraph{Comparison} We take LLaVA-v.1.5-7B as the base policy model, and compare \ours with 4 previous methods that apply preference optimization with self-generated candidate responses. These methods primarily vary in the source of reward signals: LLaVA-RLHF~\citep{sun2023aligning} leverages a pretrained reward model based on human feedback; SIMA~\citep{wang2024enhancing} develops an in-context self-critic prompt for providing pairwise judgments; CSR~\citep{zhou2024calibrated} incorporates sentence-level beam search with CLIP-score calibration; and RLAIF-V~\citep{yu2024rlaif} adopts a divide-and-conquer strategy to calculate the overall reward score by combining sentence-level judgments. For our method, we utilize the prompts (question-image pairs) from the LLaVA-RLHF dataset and perform DPO training for 3 epoches.

As illustrated in Table~\ref{tab:dpo_llava15}, with only 9.4k input prompts, the reward signal provided by \ours substantially improve the base model's performance across various open-ended visual chat benchmarks. It achieves the best improvements of +10.1 on LLaVA-W, +3.0 on LLaVA-Wilder, +8.8 on WildVision-Bench, along with the second-highest gains of + 4.4 on LiveBench and +0.13 on MMHal-Bench, respectively. At the same time, the overall capacities of LLaVA-v1.5-7B are largely preserved, as demonstrated on other comprehensive benchmarks. This is superior to other competing methods, which either result in smaller performance gains or achieve improvements by compromising the overall capabilities on other benchmarks.

\begin{table}[!tp]
    \centering
    \vspace{-3mm}
    \small
    \begin{tabular}{c|cc}
    \toprule
       Sampling  & LLaVA-W & L-Wilder \\
    \midrule
       Random  & 100.3 & 71.6\\
       Best-of-5 {\scriptsize(w. Critic-7B)} & 102.0 & 74.8 \\
    \bottomrule
    \end{tabular}
    \vspace{-1.5mm}
    \caption{Results of BoN sampling. Responses are generated by the OV-7B checkpoint after 3-round iterative DPO training, with \ours providing the reward scores.}
    \vspace{-5mm}
    \label{tab:critic_bon}
\end{table}

\vspace{-3mm}
\paragraph{Inference Time Search} Applying \ours for best-of-n sampling~\citep{stiennon2020learning} further enhances LMM performance during inference. For the LLaVA-OV-7B checkpoint after iterative DPO training, we generate $n=5$ responses for each question with a temperature of 0.7 and top-p of 0.9, then use \ours-7B to select the best responses. As shown in Table~\ref{tab:critic_bon}, this results in additional gains of +1.7 on LLaVA-W and +3.2 on LLaVA-Wilder.

\section{Conclusions}
We have presented LLaVA-Critic, an open-source LMM that is trained to evaluate model performance in a wide range of multimodal scenarios. To achieve this, we curated a high-quality critic instruction-following dataset with diverse evaluation criteria. We demonstrated the effectiveness of LLaVA-Critic in two key areas: (1) as a generalized evaluator, \ours provides pointwise scores and pairwise rankings that closely align with human and GPT-4o preferences across multiple evaluation tasks, presenting a viable open-source alternative to commercial GPT models for autonomous assessment of open-ended LMM responses; (2) in preference learning, \ours functions as a reliable reward model, supplying preference signals that enhance the visual chat capabilities of LMMs, surpassing the LLaVA-RLHF reward model built with human feedback.
This work represents an important step toward harnessing the self-critique capabilities of open-source LMMs, and we hope it will inspire further research into developing strong LMMs with scalable and superhuman alignment feedback.

{
    \small
    \bibliographystyle{ieeenat_fullname}
    \bibliography{main}
}

\appendix
\clearpage

\section{Author Contributions}
\begin{itemize}[leftmargin=7pt, label=-, itemsep=3pt]
    \item Tianyi Xiong primarily executes the \ours project. He collects pointwise samples for the \ours-113k training dataset, trains the models, implements and experiments with iterative DPO, and conducts evaluation of \ours across various scenarios. 
    \item Xiyao Wang contributes to developing pairwise ranking capacity, including organizing response pairs and preferences, designing pairwise prompt templates, and building in-domain pairwise evaluation benchmarks.
    \item Dong Guo consistently provides technical support throughout the project and contributes to launching ablation study experiments.
    \item Qinghao Ye and Haoqi Fan offer helpful suggestions and efforts for constructing the critic training dataset.
    \item Quanquan Gu gives expert guidance on preference learning and helps refine the related sections of the paper.
    \item Heng Huang provides support for the project and helps with revising the paper.
    \item Chunyuan Li initiates the idea and scope, directs the project, outlines the roadmap and milestones, supervises the experimental execution, and leads the paper writing.
\end{itemize}
\vspace{1.5mm}

\noindent We would like to extend our gratitude to Renrui Zhang for his insightful discussions and constructive feedback on the evaluation scope of \ours. Special thanks also go to Bo Li for his significant support with LLaVA-OneVision training, SGLang inference, and contributions to the early version of the DPO codebase.
\section{Implementation Details}

\subsection{Evaluation Prompts For Critic Training}
\label{append:training_prompts}

\paragraph{Pointwise prompts} To construct pointwise critic training data, we adopt the existing evaluation prompts in 7 widely used multimodal evaluation benchmarks that employ GPT-as-a-judge. For further details, please refer to their papers or codebases as listed below:

\vspace{1.5mm}
\begin{itemize}[leftmargin=10pt,itemsep=5pt]
    \item \textit{LLaVA-in-the-Wild}~\citep{liu2023llava}: \href{https://arxiv.org/abs/2304.08485}{arxiv.org/abs/2304.08485}
    \item \textit{LLaVA-Wilder}~\citep{li2024llavanext-strong}: \href{https://github.com/EvolvingLMMs-Lab/lmms-eval/blob/main/lmms_eval/tasks/llava_wilder/utils.py}{lmms\_eval/tasks/llava\_wilder}
    \item \textit{ImageDC}~\citep{li2024llavanext-strong}: \href{https://github.com/EvolvingLMMs-Lab/lmms-eval/blob/main/lmms_eval/tasks/internal_eval/dc100_en_utils.py}{lmms\_eval/tasks/internal\_eval/dc100\_en}
    \item \textit{MMHal-Bench}~\citep{sun2023aligning}: \href{https://arxiv.org/abs/2309.14525}{arxiv.org/abs/2309.14525}
    \item \textit{MM-Vet}~\citep{yu2023mmvetevaluatinglargemultimodal}: \href{https://arxiv.org/abs/2308.02490}{arxiv.org/abs/2308.02490}
    \item \textit{WildVision-Bench}~\citep{lu2024wildvision}: \href{https://arxiv.org/abs/2406.11069}{arxiv.org/abs/2406.11069}
    \item \textit{RefoMB}~\citep{yu2024rlaif}: \href{https://arxiv.org/abs/2405.17220}{arxiv.org/abs/2405.17220}
\end{itemize}
\vspace{1.5mm}

\noindent LLaVA-in-the-Wild, MM-Vet, MMHal-Bench and RefoMB use text-only GPT models for evaluation. We slightly adjust their evaluation prompts to guide the evaluator to focus on visual information from the input image.

\vspace{-3mm}
\paragraph{Pairwise Prompt Pool} To fully develop \ours's capacity of ranking LMM responses pairs across diverse scenarios, with varying format requirements and evaluation criteria, we design a set of 30 pairwise evaluation prompt templates for constructing our pairwise critic training data. Two representative prompts are provided in Table~\ref{tab:example_pairwise_prompts}.

\subsection{In-domain Benchmarks for LMM-as-a-Judge}

\label{append:indomain-pointwise-eval}

\paragraph{In-domain Pointwise Scoring} To evaluate the ability of \ours in judging LMM-generated responses across varying performance levels, we select 13 off-the-shelf LMMs spanning across a wide range of visual chat capabilities, then collect their responses on 7 multimodal benchmarks. The selected response models are listed below:

\begin{quote}
GPT-4o~\citep{openai2024gpt4o}, Claude3-Sonnet~\citep{anthropic2024claude3}, LLaVA-NeXT (LLaMA-8B)~\citep{liu2024llavanext}, LLaVA-NeXT (Vicuna-7B)~\citep{liu2024llavanext}, LLaVA-OneVision-7B~\citep{li2024llavaonevision}, LLaVA-RLHF-13B~\citep{sun2023aligning}, LLaVA-v1.5-7B~\citep{liu2024improved}, LLaVA-v1.5-13B~\citep{liu2024improved}, InstructBLIP-Vicuna-7B~\citep{dai2024instructblip}, InternVL2-8B~\citep{chen2023internvl}, Phi-3-Vision-128k-Instruct~\citep{abdin2024phi}, fuyu-8B~\citep{fuyu-8b} and Qwen-VL-Chat~\citep{bai2023qwen}
\end{quote}

\section{Additional Experimental Results}

\subsection{Ablation on Data Curation}
\begin{table}[!htp]
    \centering
    \vspace{-1mm}
    \setlength{\tabcolsep}{3pt}
    \scalebox{0.81}{
    \begin{tabular}{c|ccc}
    \toprule
      Training Data   & Score$\uparrow$ & Pair {\footnotesize (w. Tie)}$\uparrow$ & Pair {\footnotesize (w.o. Tie)}$\uparrow$ \\
    \hline
      -- & 0.151  & 0.426 & 0.550 \\
      pointwise only & 0.311 & 0.543 & 0.664 \\
      pairwise only & 0.249 & 0.541 & 0.659 \\
      pointwise + pairwise (w.o. Tie) & 0.312 & 0.552 & 0.683 \\
      all & 0.314 & 0.556 & 0.689 \\
      \bottomrule
    \end{tabular}
    }
    \vspace{-2mm}
    \caption{Ablation on data curation. We start from LLaVA-OV-7B to train critic models and evaluate them on MLLM-as-a-Judge.}
    \vspace{-3.5mm}
    \label{tab:supp_data_ablation}
\end{table}

\noindent We analyze the impact of different data components in the \ours-113k dataset for critic training. Results are reported in Table~\ref{tab:supp_data_ablation}, presenting the following insights: 
1) Pointwise and pairwise critic data are mutually beneficial. 2) Pointwise only data improves pairwise ranking more than the reverse, as relative ranking lacks absolute judgments. 3) Tie pairs further enhance performance by avoiding fixation on minor quality differences.

\begin{table*}[!tp]
\centering
\begin{minipage}{1.0\textwidth}
\vspace{0mm}    
\centering
\begin{tcolorbox} 
\centering
\begin{tabular}{p{0.96\textwidth}}
\VarSty{ {\bf Prompt Template 1:} } \vspace{1mm} \\
You will be given an image and a corresponding question. \
Your task is to evaluate the responses provided, select the better response, and give reasons for your choice.
Consider the following factors when making your decision.\vspace{2mm} \\
1. \textit{Accuracy in Object Description}: Evaluate the accuracy of the descriptions concerning the objects mentioned in the ground truth answer. Responses should minimize the mention of objects not present in the ground truth answer, and inaccuracies in the description of existing objects. \\
2. \textit{Accuracy in Depicting Relationships}: Consider how accurately the relationships between objects are described compared to the ground truth answer. Rank higher the responses that least misrepresent these relationships. \\
3. \textit{Accuracy in Describing Attributes}: Assess the accuracy in the depiction of objects' attributes compared to the ground truth answer. Responses should avoid inaccuracies in describing the characteristics of the objects present. \\
4. \textit{Helpfulness}: Consider whether the generated text provides valuable insights, additional context, or relevant information that contributes positively to the user’s comprehension of the image. Assess whether the language model accurately follows any specific instructions or guidelines provided in the prompt. Evaluate the overall contribution of the response to the user experience. \\
5. \textit{Ethical Considerations}: Identify if the model gives appropriate warnings or avoids providing advice on sensitive topics, such as medical images. Ensure the model refrains from stating identification information in the image that could compromise personal privacy. Evaluate the language model’s responses for fairness in treating individuals and communities, avoiding biases. Assess for harmfulness, ensuring the avoidance of content that may potentially incite violence, be classified as NSFW (Not Safe For Work), or involve other unmentioned ethical considerations. Consider any content that could be deemed offensive, inappropriate, or ethically problematic beyond the explicitly listed criteria. \vspace{2mm} \\
The question and responses are given as follows: \\
Question: [{}] \textbackslash n
Response A: [{}] \textbackslash n
Response B: [{}] \textbackslash n  
ASSISTANT:  \\
\hrulefill \vspace{2mm} \\
\VarSty{ {\bf Prompt Template 2:} } \vspace{1mm}\\
As an expert, you are asked to evaluate two responses to the given image-based question. \
Provide a professional assessment of responses and decide which one is better. \
Support your decision with detailed reasons. Here are the question and responses: \\
Question: [{}] \textbackslash n
The first response: [{}] \textbackslash n
The second response: [{}] \textbackslash n
ASSISTANT: \\

\end{tabular}
\end{tcolorbox}
\vspace{-2mm}
\caption{Example prompt templates for constructing pairwise training data. 
The first template covers scenarios with specific \textit{user-defined criteria}, while the second is for general evaluations without additional instructions.}
\label{tab:example_pairwise_prompts}
\end{minipage}
\end{table*}

\definecolor{front-color}{HTML}{F5FFFA}
\begin{table*}[!tp]
\centering
\setlength{\tabcolsep}{1.5pt}
\vspace{1mm}
\scalebox{0.94}{
\begin{tabular}{l|ccccccc|c}
\toprule
 \multirow{2}{*}{LMM Evaluator} & \multicolumn{8}{c}{Kendall's $\tau$ ($\uparrow$)} \\
 \cmidrule{2-9}
 & ImageDC & MMVet & WildVision & LLaVA-B & LLaVA-W & L-Wilder & MMHal & Avg. \\
\midrule 
LLaVA-NeXT (LLaMA-8B) & 0.452 & 0.436 & 0.615  & 0.487& 0.503  & 0.231  & 0.590 & 0.473 \\
\rowcolor{front-color}
LLaVA-Critic (LLaVA-NEXT) & 0.787 & 0.974	& 0.846	& 0.839 & 0.923 & 0.974 & 0.923 & 0.895 \\
\hline
Qwen2-VL-7B-Instruct & 0.667 & 0.462 & 0.128 & 0.513 & 0.667 & 0.974 & 0.487 & 0.557 \\
LLaMA3.2-11B-Vision-Instruct & 0.590 & 0.718 & 0.872 & 0.667 & 0.615 & 0.615 & 0.452 & 0.647 \\
LLaVA-OV-7B  & 0.539 & 0.154 & 0.795  & 0.667& 0.641  & 0.839  & 0.590 & 0.603 \\
\rowcolor{front-color}
LLaVA-Critic-7B (v0.5)  & 0.813 & 0.897 & 0.872  & 0.846& 0.949  & 0.974  & 0.923 & 0.896 \\
\rowcolor{front-color}
LLaVA-Critic-7B  & 0.897 & 0.949 & 0.897  & 0.839& 0.923  & 0.974  & 0.897 & 0.911   \\ \hline
LLaVA-OV-72B & 0.872 & 0.795 & 0.821  & 0.667& 0.769  & 0.949  & 0.744 & 0.802 \\
\rowcolor{front-color}
LLaVA-Critic-72B & 0.949 & 0.949 & 0.949  & 0.821& 0.923  & 0.994  & 0.949 & 0.933 \\
\bottomrule
\end{tabular}
} \\ 
\vspace{-1mm}
\caption{Kendall's Tau results on in-domain pointwise scoring. \ours consistently provides model-level rankings more aligned with GPT-4o, surpassing baseline methods across all evaluation tasks.}
\label{tab:kendall_tau_in_domain_pointwise}
\end{table*}

\begin{table*}[!tbp]
\centering
\large
\renewcommand\arraystretch{1.2}
\label{tab: full result}
\setlength{\tabcolsep}{5pt}
\resizebox{0.98\linewidth}{!}{
\begin{tabular}{ll|cccccccccccccc|c}
\toprule[1.5pt]
\textbf{Settings} & \textbf{MLLM}  & COCO  & C.C. & Diff. & Graphics & Math & Text & WIT & Chart & VisIT & CC-3M & M2W & SciQA & Aes & MM-Vet & Ave. \\ \midrule
\multirow{17}{*}{\textbf{Score} ($\uparrow$)}

 & GPT-4V* & 0.454 & 0.507 & 0.458 & 0.645 & 0.606 & 0.624 & 0.579 & 0.645 & 0.620 & 0.431 & 0.185 & 0.383 & 0.401 & 0.326 & 0.490 \\
 & Gemini-Pro* & 0.262 & 0.408 & - & 0.400 & 0.228 & 0.222 & 0.418 & 0.343 & 0.336 & 0.374 & 0.324 & 0.073 & 0.360 & 0.207 & 0.304 \\
 & LLaVA-1.5-13b* & 0.247 & 0.227 & 0.060 & 0.242 & 0.093 & 0.245 & 0.109 & 0.237 & 0.177 & 0.071 & 0.424 & 0.279 & 0.414 & 0.322 & 0.225 \\
 \cline{2-17}
  \rowcolor{light-gray} \cellcolor{white}
 & GPT-4o\textsuperscript{\textdagger} & 0.396 & 0.452 & 0.341 & 0.464 & 0.460 & 0.564 & 0.408 & 0.573 & 0.589 & 0.305 & 0.262 & 0.569 & 0.421 & 0.342 & 0.439 \\
  \rowcolor{light-gray} \cellcolor{white}
 & GPT-4V\textsuperscript{\textdagger} & 0.410 & 0.444 & 0.361 & 0.449 & 0.486 & 0.506 & 0.457 & 0.585 & 0.554 & 0.266 & 0.267 & 0.315 & 0.472 & 0.367  & 0.424 \\
 & LLaVA-v1.5-7B & 0.205 & 0.084 & 0.094 & -0.028 & 0.099 & 0.235 & -0.076 & 0.212 & 0.300 & 0.167 & 0.392 & 0.033 & 0.354 & 0.147 & 0.158 \\
 & Prometheus-V (LLaVA-v1.5-7B) & 0.289 & 0.342 & 0.106 & 0.172 & 0.182 & 0.214 & 0.209 & 0.224 & 0.226 & 0.228 & 0.089 & 0.174 & 0.368 & 0.157 & 0.213 \\
 & LLaVA-NeXT (LLaMA-8B) & 0.288 & 0.279 & 0.066 & 0.249 & 0.077 & 0.145 & -0.051 & 0.197 & 0.304 & 0.198 & 0.364 & 0.128 & 0.362 & 0.165 & 0.198 \\
  & LLaVA-OV-7B &0.224 &0.024 &0.063 &0.189 &0.097 & 0.265 & -0.135 & 0.274 & 0.227 & 0.081 & 0.030 & 0.261 & 0.249 & 0.262 & 0.151 \\
  & Qwen2-VL-7B-Instruct & 0.295 & 0.294 & 0.114 & 0.196 & 0.199 & 0.323 & 0.075 & 0.366 & 0.259 & 0.143 & 0.313 & 0.201 & 0.434 & 0.331 & 0.253 \\
   & LLaMA3.2-11B-Vision-Instruct & 0.269 & 0.244 & 0.107 & 0.365 & 0.254 & 0.305 & 0.024 & 0.333 & 0.264 & 0.067 & 0.197 & 0.370 & 0.275 & 0.245 & 0.237 \\
  & LLaVA-OV-72B & 0.264 & 0.390 & 0.046 & 0.262 & 0.358 & 0.327 & 0.195 & 0.290 & 0.415 & 0.144 & 0.359 & 0.267 & 0.444 & 0.253 &  0.287 \\  
   \rowcolor{front-color}
  \cellcolor{white} & \ours (LLaVA-v1.5-7B) & 0.283 & 0.295 & 0.095 & 0.225 & 0.246 & 0.257 & 0.191 & 0.194 & 0.253 & 0.241 & 0.198 & 0.188 & 0.327 & 0.198 & 0.228 \\
   \rowcolor{front-color}
  \cellcolor{white} & \ours (LLaVA-NeXT) & 0.272 & 0.406 & 0.118 & 0.257 & 0.309 & 0.275 & 0.292 & 0.354 & 0.374 & 0.225 & 0.224 & 0.091 & 0.432 & 0.176 & 0.272 \\
 \rowcolor{front-color}
  \cellcolor{white} & \ours-7B (v0.5) & 0.369 & 0.456 & 0.108 & 0.291 & 0.325 & 0.340 & 0.228 & 0.388 & 0.303 & 0.205 & 0.232 & 0.270 & 0.511 & 0.338 & 0.312 \\
  \rowcolor{front-color}
  \cellcolor{white} & \ours-7B & 0.382 & 0.450 & 0.103 & 0.316 & 0.356 & 0.378 & 0.179 & 0.421 & 0.322 & 0.246 & 0.301 & 0.269 & 0.395 & 0.272 & 0.314 \\
   \rowcolor{front-color}
  \cellcolor{white} & \ours-72B & 0.333 & 0.463 & 0.146 & 0.452 & 0.474 & 0.559 & 0.396 & 0.545 & 0.488 & 0.273 & 0.259 & 0.334 & 0.403 & 0.374 & 0.393 \\
\midrule
\multirow{16}{*}{\textbf{Pair w. Tie} ($\uparrow$)} 
 & GPT-4V* & 0.696 & 0.824 & 0.847 & 0.639 & 0.564 & 0.673 & 0.679 & 0.657 & 0.640 & 0.612 & 0.521 & 0.415 & 0.606 & 0.529 & 0.636 \\
 & Gemini-Pro* & 0.616 & 0.787 & - & 0.650 & 0.436 & 0.664 & 0.605 & 0.500 & 0.660 & 0.560 & 0.370 & 0.262 & 0.190 & 0.312 & 0.509 \\
 & Qwen-vl-plus* & 0.479 & 0.507 & 0.650 & 0.450 & 0.328 & 0.522 & 0.500 & 0.380 & 0.453 & 0.383 & 0.577 & 0.321 & 0.601 & 0.457 & 0.472 \\
 \cline{2-17}
 \rowcolor{light-gray} \cellcolor{white}
 & GPT-4o\textsuperscript{\textdagger} & 0.582 & 0.665 & 0.829 & 0.625 & 0.433 & 0.477 & 0.565 & 0.355 & 0.577 & 0.586 & 0.581 & 0.427 & 0.873 & 0.505 & 0.577 \\
  \rowcolor{light-gray} \cellcolor{white}
 & GPT-4V\textsuperscript{\textdagger} & 0.539 & 0.634 & 0.668 & 0.632 & 0.459 & 0.495 & 0.536 & 0.369 & 0.591 & 0.544 & 0.544 & 0.389 & 0.620 & 0.517 & 0.538 \\
  & LLaVA-v1.5-7B & 0.460 & 0.506 & 0.577 & 0.452 & 0.302 & 0.454 & 0.375 & 0.383 & 0.518 & 0.478 & 0.495 & 0.281 & 0.466 & 0.402 & 0.439 \\
  & LLaVA-NeXT (LLaMA-8B) & 0.422 & 0.595 & 0.435 & 0.404 & 0.343 & 0.431 & 0.428 & 0.384 & 0.508 & 0.496 & 0.571 & 0.336 & 0.588 & 0.512 & 0.461 \\
  & LLaVA-OV-7B & 0.334 & 0.471 & 0.539 & 0.397 & 0.318 & 0.398 & 0.324 & 0.374 & 0.444 & 0.438 & 0.556 & 0.334 & 0.577 & 0.456 & 0.426\\
  & Qwen2-VL-7B-Instruct & 0.391 & 0.358 & 0.213 & 0.413 & 0.368 & 0.383 & 0.231 & 0.308 & 0.391 & 0.406 & 0.324 & 0.438 & 0.281 & 0.367 & 0.348 \\
   & LLaMA3.2-11B-Vision-Instruct & 0.493 & 0.643 & 0.638 & 0.601 & 0.422 & 0.512 & 0.485 & 0.384 & 0.544 & 0.591 & 0.579 & 0.380 & 0.662 & 0.466 & 0.529 \\
   & LLaVA-OV-72B & 0.464 & 0.593 & 0.667 & 0.531 & 0.434 & 0.485 & 0.447 & 0.394 & 0.549 & 0.497 & 0.557 & 0.428 & 0.596 & 0.541 & 0.513 \\
 \rowcolor{front-color}
  \cellcolor{white} & \ours (LLaVA-v1.5-7B) & 0.564 & 0.674 & 0.633 & 0.505 & 0.422 & 0.528 & 0.538 & 0.386 & 0.583 & 0.608 & 0.577 & 0.294 & 0.681 & 0.404 & 0.528 \\
\rowcolor{front-color}
  \cellcolor{white} & \ours (LLaVA-NeXT) & 0.583 & 0.684 & 0.704 & 0.562 & 0.438 & 0.504 & 0.579 & 0.339 & 0.635 & 0.599 & 0.581 & 0.315 & 0.693 & 0.441 & 0.547 \\
 \rowcolor{front-color}
  \cellcolor{white} & \ours-7B (v0.5) & 0.575 & 0.677 & 0.730 & 0.556 & 0.427 & 0.521 & 0.537 & 0.366 & 0.568 & 0.620 & 0.571 & 0.353 & 0.703 & 0.435 & 0.546 \\
\rowcolor{front-color}
  \cellcolor{white} & \ours-7B & 0.593 & 0.687 & 0.707 & 0.587 & 0.432 & 0.544 & 0.564 & 0.338 & 0.596 & 0.628 & 0.591 & 0.370 & 0.686 & 0.464 & 0.556 \\
\rowcolor{front-color}
  \cellcolor{white} & \ours-72B & 0.587 & 0.672 & 0.860 & 0.588 & 0.475 & 0.536 & 0.618 & 0.366 & 0.628 & 0.608 & 0.568 & 0.390 & 0.721 & 0.473 & 0.578 \\
 \midrule
\multirow{16}{*}{\textbf{Pair w.o. Tie} ($\uparrow$)} 

 & GPT-4V* & 0.804 & 0.870 & 0.922 & 0.807 & 0.801 & 0.805 & 0.734 & 0.849 & 0.761 & 0.703 & 0.699 & 0.647 & 0.755 & 0.659 & 0.773 \\
 & Gemini-Pro* & 0.717 & 0.840 & - & 0.770 & 0.678 & 0.793 & 0.688 & 0.658 & 0.711 & 0.652 & 0.471 & 0.358 & 0.265 & 0.400 & 0.615 \\
 & LLaVA-1.6-34b* & 0.607 & 0.824 & 0.855 & 0.402 & 0.587 & 0.750 & 0.758 & 0.381 & 0.503 & 0.564 & 0.712 & 0.679 & 0.694 & 0.762 & 0.648 \\
 \cline{2-17}
  \rowcolor{light-gray} \cellcolor{white}
 & GPT-4o\textsuperscript{\textdagger} & 0.774 & 0.776 & 0.934 & 0.835 & 0.628 & 0.618 & 0.737 & 0.513 & 0.741 & 0.770 & 0.706 & 0.722 & 0.887 & 0.660 & 0.736 \\
  \rowcolor{light-gray} \cellcolor{white}
 & GPT-4V\textsuperscript{\textdagger} & 0.729 & 0.772 & 0.884 & 0.853 & 0.665 & 0.661 & 0.760 & 0.495 & 0.785 & 0.707 & 0.697 & 0.639 & 0.741 & 0.654 & 0.717 \\
 & LLaVA-v1.5-7B & 0.617 & 0.571 & 0.637 & 0.598 & 0.411 & 0.544 & 0.452 & 0.554 & 0.653 & 0.562 & 0.672 & 0.600 & 0.558 & 0.631 & 0.576 \\
 & LLaVA-NeXT (LLaMA-8B) & 0.565 & 0.684 & 0.473 & 0.526 & 0.460 & 0.526 & 0.516 & 0.549 & 0.634 & 0.592 & 0.641 & 0.648 & 0.673 & 0.716 & 0.586 \\
  & LLaVA-OV-7B &  0.462 & 0.562 & 0.588 & 0.530 & 0.434 & 0.473 & 0.400 & 0.543 & 0.563 & 0.527 & 0.639 & 0.670 & 0.633 & 0.679 & 0.550 \\
  & Qwen2-VL-7B-Instruct & 0.662 & 0.797 & 0.579 & 0.753 & 0.576 & 0.600 & 0.418 & 0.444 & 0.735 & 0.712 & 0.712 & 0.642 & 0.848 & 0.550 & 0.645 \\
  & LLaMA3.2-11B-Vision-Instruct & 0.647 & 0.726 & 0.688 & 0.774 & 0.570 & 0.599 & 0.577 & 0.567 & 0.667 & 0.692 & 0.654 & 0.708 & 0.684 & 0.662 & 0.658 \\
   & LLaVA-OV-72B & 0.691 & 0.780 & 0.811 & 0.714 & 0.623 & 0.634 & 0.625 & 0.570 & 0.737 & 0.685 & 0.755 & 0.702 & 0.746 & 0.736 & 0.701 \\

 \rowcolor{front-color}
  \cellcolor{white} & \ours (LLaVA-v1.5-7B) & 0.732 & 0.757 & 0.665 & 0.659 & 0.574 & 0.623 & 0.650 & 0.544 & 0.710 & 0.719 & 0.649 & 0.589 & 0.708 & 0.602 & 0.656 \\
  \rowcolor{front-color}
  \cellcolor{white} & \ours (LLaVA-NeXT) & 0.763 & 0.775 & 0.762 & 0.720 & 0.599 & 0.606 & 0.705 & 0.491 & 0.756 & 0.716 & 0.682 & 0.598 & 0.725 & 0.577 & 0.677\\
   \rowcolor{front-color}
  \cellcolor{white} & \ours-7B (v0.5) & 0.747 & 0.758 & 0.771 & 0.716 & 0.580 & 0.625 & 0.661 & 0.525 & 0.692 & 0.729 & 0.697 & 0.632 & 0.728 & 0.585 & 0.675\\
  \rowcolor{front-color}
  \rowcolor{front-color}
  \cellcolor{white} & \ours-7B & 0.771 & 0.774 & 0.755 & 0.758 & 0.596 & 0.658 & 0.680 & 0.488 & 0.727 & 0.742 & 0.692 & 0.658 & 0.715 & 0.635 & 0.689 \\
 \rowcolor{front-color}
  \cellcolor{white} & \ours-72B & 0.762 & 0.762 & 0.904 & 0.755 & 0.637 & 0.648 & 0.763 & 0.528 & 0.769 & 0.718 & 0.693 & 0.708 & 0.742 & 0.624 & 0.715 \\
 \bottomrule[1.5pt]
\end{tabular}}

\caption{Comprehensive results on MLLM-as-a-Judge. *: the results of GPT-4V, Gemini-pro, and the best open-source LMM as reported in the original paper~\citep{chen2024mllm}; \textsuperscript{\textdagger}: results from our evaluation of GPT-4V/4o using its original codebase.
}
\label{tab:mllm_judge_comprehensive}
  \vspace{3mm}
\end{table*}

\begin{table*}[!ht]
    \centering
    \setlength{\tabcolsep}{4pt}
    \scalebox{0.92}{
    \begin{tabular}{l|ccccc|ccccc}
    \toprule
      Method   & LLaVA-W & L-Wilder & WildVision & LiveBench &  MMHal &  $\text{MME}^{P}$ & $\text{MME}^{C}$ & MMB-en & MM-Vet & MMStar \\
    \midrule
      LLaVA-OV-7B  & 90.7 & 67.8 & 54.0 & 77.1 &  3.19 & 1580.4 & 418.2 & 80.8 & 57.5 & 61.7\\
      \ours-7B & 90.9 & 69.4 & 54.5 & 76.2 & 3.45 & 1599.0 & 455.4 & 79.3 & 58.4 & 58.9 \\
    \midrule
     LLaVA-OV-72B & 93.5 & 72.0 & 51.7 & 81.5 & 3.61 & 1683.2 & 578.9 & 85.9 & 63.7 & 66.1 \\
      \ours-72B & 93.8 &  73.4 & 54.8  & 81.9 & 3.74 & 1726.7 & 597.1 & 84.8 & 62.3 & 64.2 \\
    \bottomrule
    \end{tabular}
    }
    \vspace{-1mm}
    \caption{Comparison between \ours and LLaVA-OneVision on multimodal evaluation benchmarks. Results are reported for 5 open-ended visual chat benchmarks (left) and 4 comprehensive benchmarks (right).} 
    \label{tab:critic_comprehensive}
\end{table*}

\begin{table*}[!ht]
    \centering
    \small
    \setlength{\tabcolsep}{1.2pt}
    \scalebox{0.96}{
    \begin{tabular}{l|cccccc|ccccc}
    \toprule
      Method   & LLaVA-W & L-Wilder & WildVision & LiveBench & VideoDC & MMHal &  $\text{MME}^{P}$ & $\text{MME}^{C}$ & MMB-en & MM-Vet & MMStar \\
    \midrule
    \rowcolor{light-gray} GPT-4V & 98.0 & 81.0 & 79.8 & 73.7 & 4.00 & 3.83 & 1409.4 & 517.1 & 75.0 & 49.9 & 57.1 \\
    \midrule
      LLaVA-OV-7B  & 90.7 & 67.8 & 54.0 & 77.1 &  3.75 & 3.19 & \textbf{1580.4} & 418.2 & \textbf{80.8} & \textbf{57.5} & 61.7\\
      + \ours-7B iter-1 & 96.7 &  70.6 & 60.5 & 81.2 & 3.77 & 3.62 & 1561.8 & \textbf{420.7} & \textbf{80.8} & 54.5 & 62.1 \\
      + \ours-7B iter-2 & 97.0 &  \textbf{72.2} & 65.2 & 83.9 & 3.82 & 3.67 & 1565.5 & 415.4 & 80.7  & 54.6 & 62.1\\
      + \ours-7B iter-3 & \textbf{100.3} & 71.6 & \textbf{67.3} & \textbf{84.5} & \textbf{3.87} & \textbf{3.91} & 1555.3 & 414.6 & 80.3 & 54.4 & \textbf{62.3} \\
      \rowcolor{light-gray} + GPT-4o iter-3 & 100.6 & 71.3 & 67.1 & 86.4 & -- & 3.79 & 1565.3 & 405.0 & 80.3 & 54.8 & 59.0 \\
    \midrule
     LLaVA-OV-72B & 93.5 & 72.0 & 51.7 & 81.5 & 3.60 & 3.61 & 1683.2 & 578.9 & \textbf{85.9} & 63.7 & 66.1 \\
      + \ours-72B iter-1 & 99.3 & 75.3 & 65.7 & 86.4 & 3.83 & 3.75 & 1683.2 & 584.3 & 85.6 & 67.0 & 66.4 \\
      + \ours-72B iter-2 & 104.1 & 75.6 & 68.4 & 86.6 & \textbf{3.86} & 3.75 & 1681.1 & \textbf{586.4}  & 85.7 & 66.7 & \textbf{66.5} \\
      + \ours-72B iter-3 & \textbf{104.4} & \textbf{75.9} & \textbf{70.0} & \textbf{88.5} & \textbf{3.86} & \textbf{3.77}  & \textbf{1686.1} & \textbf{586.4} & 85.4 & \textbf{67.1} & 66.4 \\
    \bottomrule
    \end{tabular}
    }
    \vspace{-1.5mm}
    \caption{Performance of difference rounds of iterative DPO on LLaVA-OneVision. With the high-quality feedback from \ours, both LLaVA-OneVision 7B and 72B learn to refine its self-generated responses in a progressive manner, leading to overall better performance across various open-ended multimodal benchmarks.}
    \label{tab:dpo_ov7b-iter}
    \vspace{-4.5mm}
\end{table*}

\subsection{Results on In-domain Pointwise Scoring}
\label{append:pointwise-kendall-tau}
In Table~\ref{tab:kendall_tau_in_domain_pointwise}, we report the Kendall's Tau results to assess the alignment between \ours and GPT-4o in ranking response LMMs. LLaVA-Critic-72B achieves the highest average Kendall's Tau of 0.933, outperforming the LLaVA-OV-72B baseline (0.802).
\ours-7B scores 0.911, significantly surpassing existing advanced LMMs of similar scale. As with the Pearson-r results in Table~\ref{tab:result_in_domain_pointwise}, \ours benefits from both model and data scaling.

\subsection{Results on MLLM-as-a-Judge}
\label{apppend:mllm-as-a-judge-results}
We include the comprehensive results on MLLM-as-a-Judge~\citep{chen2024mllm} in Table~\ref{tab:mllm_judge_comprehensive}. Since the complete evaluation pipeline has not been released, we re-implement the evaluation code based on their official codebase.

\subsection{Results on General Visual Tasks}
\label{append:critic_visual}
We evaluate \ours on comprehensive multimodal benchmarks, including both short-answer and open-ended visual tasks.
As shown in Table~\ref{tab:critic_comprehensive}, training on critic instruction-following data effectively preserves the original capabilities for handling a wide range of visual tasks, with slight improvements in visual chat performance. The \ours-7B model outperforms LLaVA-OV-7B on 4 of 5 open-ended visual chat benchmarks, while the Critic-72B model outperforms OV-72B on all 5.

\begin{figure}[!ht]
    \centering
    \vspace{-1.5mm}
    \includegraphics[width=1.0\linewidth]{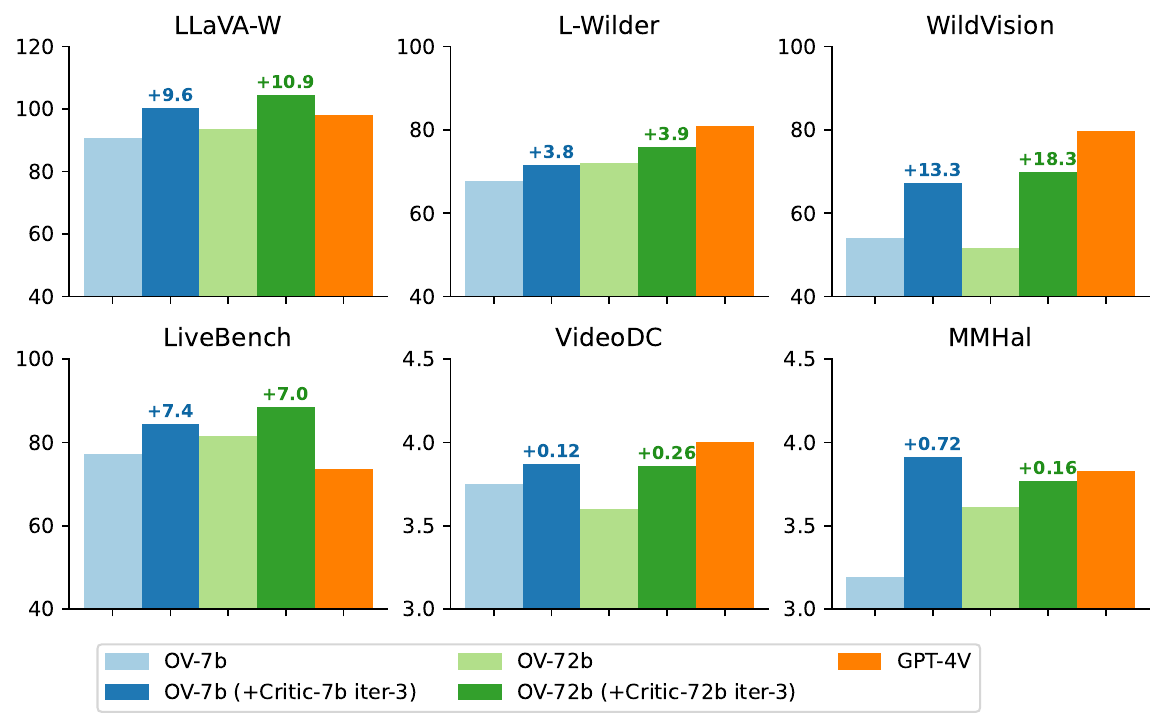}
    \caption{Visualization on the performance gain of LLaVA-OneVision obtained through \ours across 6 open-ended multimodal benchmarks. }
    \vspace{-2mm}
    \label{fig:vis_dpo_chat_gain}

\end{figure}

\subsection{Results on Preference Learning}
\label{append:dpo_results}

\noindent We present the comprehensive results of applying multi-round iterative DPO on the base model, using LLAVA-Critic for reward signals. As shown in Table~\ref{tab:dpo_ov7b-iter}, this approach enhances LLaVA-OneVision's performance across six open-ended visual chat benchmarks, with consistent gains per round. Notably, on the LLaVA-OV-7B base model, LLaVA-Critic boosts improvement comparable to GPT-4o while saving approximately \$690 in 3 iterative DPO rounds. For other comprehensive benchmarks, the iterations offer minimal benefit, while LLaVA-OneVision's overall capabilities are largely preserved. Figure~\ref{fig:vis_dpo_chat_gain} visualizes the performance gains on visual chat benchmarks.


\section{More Qualitative Comparisons}

\label{append:more_examples}

We provide detailed examples of \ours in serving as a generalist evaluator for LMM responses, and improving visual chat through preference alignment.

\begin{itemize}[leftmargin=15pt,itemsep=5pt]
    \item Table~\ref{tab:critic_example_seven} shows an example of \ours ranking a pair of LMM responses. By accurately recognizing the visual content of the input image and grounding the differences between the responses, \ours offers judgments consistent with human evaluators, along with clear justifications.

    \item Table~\ref{tab:critic_example_beach} presents an example of \ours scoring an LMM response using a reference answer from GPT-4. \ours closely follows the evaluation prompt and, by referring to the image content, accurately identifies the strengths and weaknesses of the response at both overall and fine-grained levels. Consequently, it delivers a fair evaluation score and detailed justification aligned with GPT-4o.

    \item Table~\ref{tab:example_chat_dpo} illustrates \ours's effectiveness in enhancing LLaVA-OneVision's visual chat performance through iterative DPO training. By progressively aligning with \ours's preferences, LLaVA-OneVision learns to generate responses that are more structured, detailed and better tailored to user needs.

\end{itemize}

\setlength{\fboxrule}{1pt}
\begin{table*}[!htbp]
\begin{minipage}{1.0\textwidth}
\centering
\scalebox{0.98}{
\begin{tabular}{l p{13cm}}
\toprule
 \multicolumn{2}{l}{\bf Example 1: \ours in Judging LMM Responses (Pairwise)}  \\
\midrule
&  \fbox{\includegraphics[height=2.7cm]{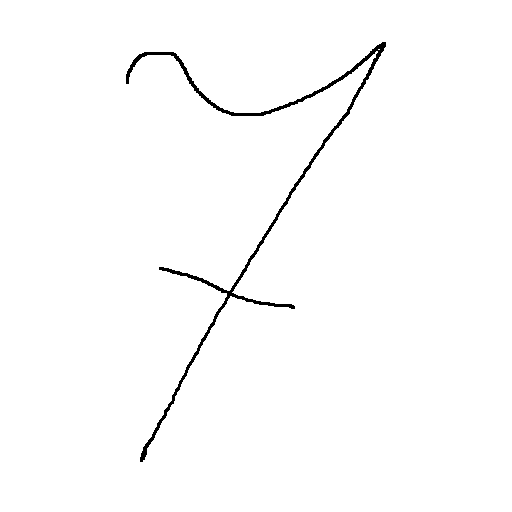}}\vspace{3pt} \\
Critic Input & \textbf{Question:} What this image presents? \\
& \textbf{Response 1 (model\_a)}: The image is a black and white sketch of a line that appears to be in the shape of a cross. The line is a simple and straightforward representation of the cross shape, with two straight lines intersecting at a point. \\
& \textbf{Response 2 (model\_b)}: This is a handwritten number seven. \\
\rowcolor{lightpurple} \cellcolor{white} & \textbf{System}: Given an image and a corresponding question, please serve as an unbiased and fair judge to evaluate the quality of the answers provided by a Large Multimodal Model (LMM). Determine which answer is better and explain your reasoning with specific details. Your task is provided as follows:\\
\rowcolor{lightpurple} \cellcolor{white} & Question: []\textbackslash n
Response 1: []\textbackslash n
Response 2: [{}]\textbackslash n
ASSISTANT: \\
\midrule
Human  & The winner is Response 2. \\
\midrule
\ours-7B & \textcolor{red}{Response 2 is better than Response 1} because it accurately identifies the image as a handwritten number seven. Response 1 incorrectly describes the image as a black and white sketch of a line in the shape of a cross, which does not match the visual content of the image. Response 2 correctly interprets the image, providing a precise and accurate description that aligns with the actual content of the image. This makes Response 2 more helpful and accurate for understanding what the image presents. \\
 \midrule
LLaVA-OV-7B & \textcolor{brown}{Response 1 is better} because it accurately describes the image as a simple and straightforward representation of a cross shape, which is what the image actually depicts. Response 2, while also describing the image, does not accurately convey the intended meaning of the image, as it suggests that the image is a handwritten number seven, which is not the case. \\
\bottomrule
\end{tabular}
}
\vspace{0mm}
\captionof{table}{Example of \ours in ranking LMM response pairs. While LLaVA-OV misinterprets the handwritten number and gives \textcolor{brown}{inaccurate judgment}, \ours \textcolor{red}{aligns with human assessments} and offers clear, visually grounded justifications.}
\label{tab:critic_example_seven}  
\end{minipage}
\end{table*}

\begin{table*}[!htbp]
\begin{minipage}{1.0\textwidth}
\centering
\footnotesize
\scalebox{0.99}{
\begin{tabular}{l p{13.5cm}}
\toprule
 \multicolumn{2}{l}{\bf Example 2: \ours in Judging LMM Responses (Pointwise)}  \\
\midrule
&  \includegraphics[height=2.7cm]{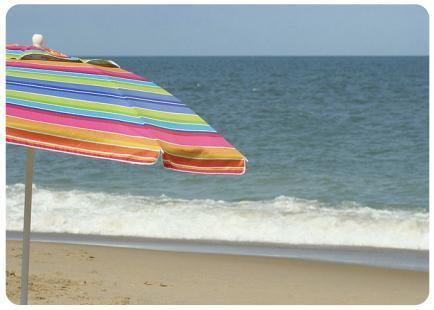} \\

Critic Input & \textbf{Question:} Can you describe the weather conditions and overall atmosphere of the scene depicted in the image? \\
& \textbf{Assistant 1 (GPT-4)}: The image depicts a calm and serene beach scene. The sky appears to be clear with no signs of clouds, indicating good weather conditions. The calmness of the ocean, suggested by the gentle waves crashing on the shore, further enhances the tranquility. The presence of a colorful beach umbrella provides a vibrant contrast to the more subdued hues of the sea and sand, implying a relaxed and leisurely atmosphere commonly associated with a sunny day at the beach. \\
& \textbf{Assistant 2 (Evaluated LMM response)}: The image depicts a sunny and pleasant day at the beach, with a colorful umbrella providing shade on the sandy shore. The presence of the umbrella suggests that it is likely a hot day, and people might be seeking relief from the sun by sitting under the umbrella or going for a swim in the ocean. The atmosphere is likely relaxed and enjoyable, with people spending time at the beach to unwind and take in the beautiful surroundings. The sunny weather and the calming sound of the waves further contribute to the overall pleasant and leisurely vibe of the scene. \\
\rowcolor{lightpurple} \cellcolor{white} & \textbf{System}: We would like to request your feedback on the performance of two AI assistants in response to the user question displayed above. The user asks the question on observing an image shown to you. Please rate the helpfulness, relevance, accuracy, level of details of their responses. Each assistant receives an overall score on a scale of 1 to 10, where a higher score indicates better overall performance. Assume assistant 1 always receive a score of 10 and is the correct answer. Please first output a single line containing only two values indicating the scores for Assistant 1 and 2, respectively. The two scores are separated by a space. In the subsequent line, please provide a comprehensive explanation of your evaluation, avoiding any potential bias and ensuring that the order in which the responses were presented does not affect your judgment. \\
\midrule
GPT-4o  & 10 9 \\
 &  Assistant 1 provided a detailed and accurate description of the weather conditions and overall atmosphere of the scene depicted in the image. The response \textcolor{blue}{highlighted the clear sky, calm ocean, and the presence of a colorful beach umbrella, which all contribute to a serene and tranquil atmosphere.} The description was comprehensive and captured the essence of the scene effectively. Assistant 2 also provided a good description, \textcolor{blue}{noting the sunny and pleasant weather, the colorful umbrella, and the relaxed atmosphere.} However, the response \textcolor{red}{included some assumptions, such as people seeking relief from the sun and going for a swim, which were not directly observable in the image.} This slightly reduced the accuracy and relevance of the response. \\
\midrule
\ours-7B & 10 9 \\
 & Assistant 1 provided a detailed and accurate description of the weather conditions and overall atmosphere depicted in the image. The response \textcolor{blue}{highlighted the clear sky, calm ocean, and the presence of a colorful beach umbrella, which all contribute to a serene and tranquil beach scene.} The description was relevant, accurate, and included a good level of detail. Assistant 2 also gave a good description, \textcolor{blue}{mentioning the sunny day, the colorful umbrella, and the relaxed atmosphere}. However, \textcolor{red}{it made an assumption about people seeking relief from the sun and going for a swim, which is not directly supported by the image. This slightly detracts from the accuracy and relevance of the response.} Therefore, Assistant 2 receives a slightly lower score. \\
 \midrule
LLaVA-OV-7B &  10 9 \\
 & Assistant 1 provided a clear and detailed description of the image, including the weather conditions and overall atmosphere. \textcolor{brown}{The response was relevant to the question asked and accurately described the scene depicted in the image. The level of detail was appropriate for the task, providing enough information for the user to understand the context of the image.} Assistant 2 also provided a clear and detailed description of the image, but it was slightly less concise than Assistant 1's response. \textcolor{brown}{While the response was still relevant to the question asked and accurately described the scene depicted in the image, it did not provide as much detail as Assistant 1's response. However, the level of detail provided by Assistant 2 was still sufficient for the task, and the response was still helpful and accurate.} Overall, both assistants performed well in their responses, but Assistant 1 provided a slightly more detailed and comprehensive response, which is why it receives a score of 10, while Assistant 2 receives a score of 9. \\
\bottomrule
\end{tabular}
}
\vspace{0mm}
\captionof{table}{Example of \ours conducting pointwise scoring on an LMM response for LLaVA-Wilder. Unlike LLaVA-OV, which offers \textcolor{brown}{vague and uninformative explanations}, \ours delivers a thorough and fair assessment with clear justifications on both the \textcolor{blue}{strengths} and \textcolor{red}{weaknesses} of the evaluated response, closely resembling the depth and fairness of GPT-4o's assessments.}
\label{tab:critic_example_beach}  
\end{minipage}
\end{table*}

\begin{table*}[!htbp]
\begin{minipage}{1.0\textwidth}
\centering
\scalebox{0.98}{
    \begin{tabular}{l p{13cm}}
    \toprule
     \multicolumn{2}{l}{\bf Example 3: \ours Enhances LLaVA-OneVision in Visual Chat}  \\
    \midrule
     & \includegraphics[height=2.7cm]{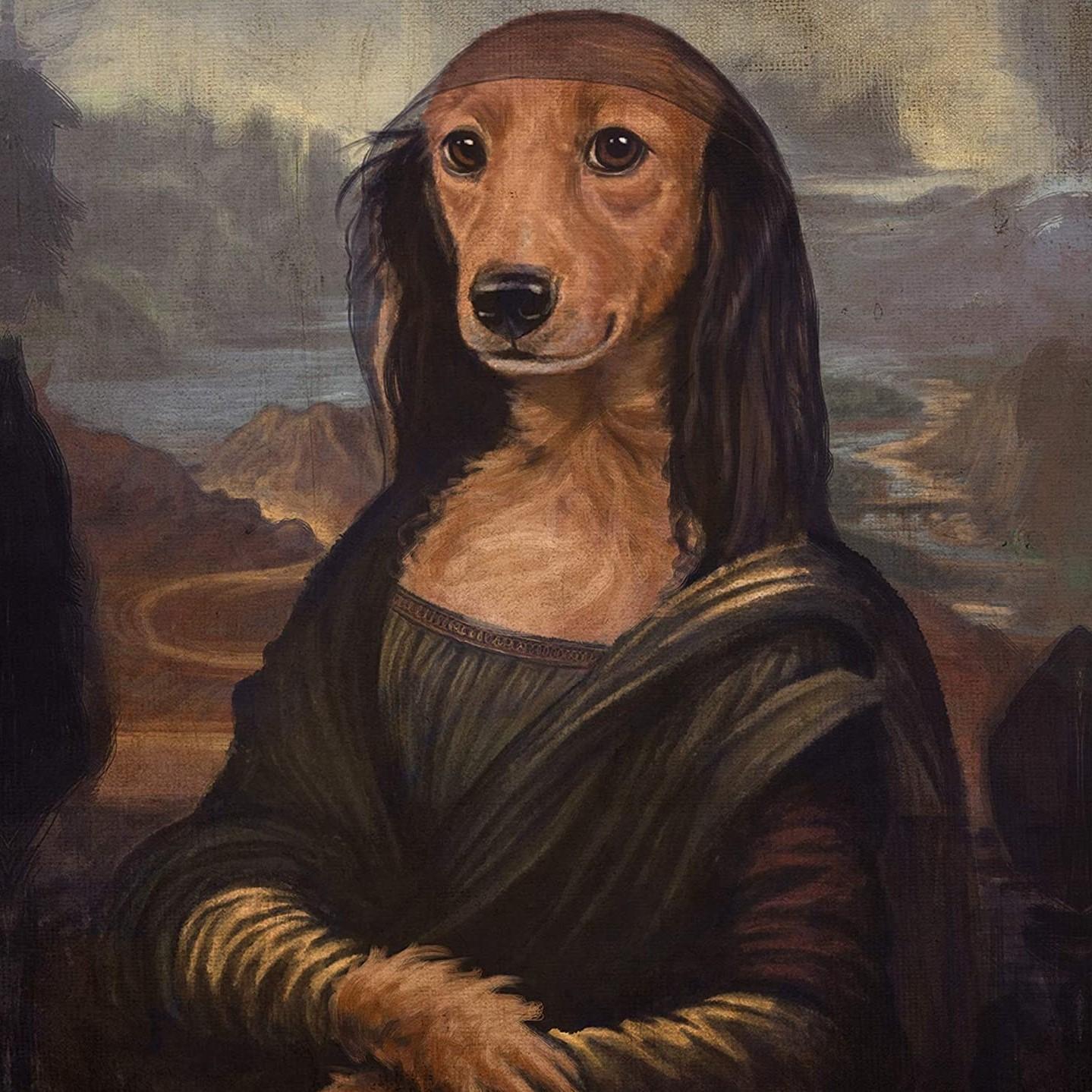} \\
    Question & Discuss how this creative twist on a classic work of art might be interpreted differently by various audiences. \\
    \midrule
    LLaVA-OV-72B & Different audiences might interpret this creative twist on the Mona Lisa in various ways. Art enthusiasts might appreciate the skillful execution and the playful reinterpretation of a classic, seeing it as a commentary on the universality of beauty and expression across species. Dog lovers could find joy in seeing their favorite breed portrayed with such dignity and elegance. Some might view it as a humorous take on high art, making it more accessible and relatable. Others might see it as a statement on the anthropomorphism of pets, reflecting how we often attribute human-like qualities to animals. Overall, the varied interpretations highlight the versatility of art in engaging diverse perspectives and emotions. \\
    \midrule
    \makecell[lt]{LLaVA-OV-72B \\ + Critic-72b iter-3} & This creative twist on the Mona Lisa, featuring a dog's face, can be interpreted in various ways by different audiences. \vspace{1mm}\\
    & For art enthusiasts, it might be seen as a playful commentary on the nature of portraiture and the human tendency to anthropomorphize animals. It challenges traditional notions of beauty and subjectivity in art, inviting viewers to reconsider what makes a portrait compelling. \vspace{1mm}\\
    & For dog lovers and pet owners, the image could evoke a sense of joy and familiarity, as they see their beloved pets reflected in one of the most iconic artworks in history. It might also highlight the deep bond between humans and their pets, suggesting that animals share in our cultural heritage. \vspace{1mm}\\
    & In a broader social context, this image could be interpreted as a reflection of contemporary society's obsession with internet culture and memes. The substitution of the Mona Lisa's face with a dog's could be seen as a satirical take on how viral content often overshadows classical art in modern media. \vspace{1mm} \\
    & Academically, this piece could spark discussions about copyright, artistic originality, and the public domain. It raises questions about the boundaries between homage and appropriation in art, encouraging critical thinking about intellectual property rights and creative freedom. \vspace{1mm}\\
    & Overall, the diverse interpretations of this image underscore the power of art to engage, provoke, and connect people across different backgrounds and perspectives. \\
    \bottomrule
    \end{tabular}
}
\captionof{table}{Example of \ours in preference learning. Leveraging reward signals from \ours, LLaVA-OneVision improves its visual chat ability through iterative DPO, delivering more detailed, valuable, and structured point-by-point responses.}
\label{tab:example_chat_dpo}  
\end{minipage}
\end{table*}

\end{document}